\documentclass[11pt]{article}
\usepackage{coling2018}
\usepackage{times}
\usepackage{url}
\usepackage{latexsym}
\usepackage{epsfig}
\usepackage{graphicx}
\usepackage{amsmath}
\usepackage{amssymb}
\usepackage{multirow}
\usepackage[font=small]{caption}
\usepackage{subfigure}
\usepackage{color}
\usepackage{epstopdf}
\usepackage{xspace}
\usepackage{pdflscape}

\usepackage{amsmath,amsfonts}
\usepackage{amssymb,amsopn}
\usepackage{bm} 

\usepackage{amssymb}
\usepackage{amsmath,amsfonts}
\usepackage{amsthm,amsopn}

\usepackage{bm} 

\newcommand{\mat}[1]{\boldsymbol{#1}} 
\newcommand{\cst}[1]{\mathsf{#1}}  

\newcommand{\ProbOpr}[1]{\mathbb{#1}}

\newcommand{\expect}[2]{%
\ifthenelse{\equal{#2}{}}{\ProbOpr{E}_{#1}}
{\ifthenelse{\equal{#1}{}}{\ProbOpr{E}\left[#2\right]}{\ProbOpr{E}_{#1}\left[#2\right]}}} 
\newcommand{\var}[2]{%
\ifthenelse{\equal{#2}{}}{\ProbOpr{VAR}_{#1}}
{\ifthenelse{\equal{#1}{}}{\ProbOpr{VAR}\left[#2\right]}{\ProbOpr{VAR}_{#1}\left[#2\right]}}} 

\newcommand{\mD}{\mat{D}}

\newcommand{\eat}[1]{}

\newcommand{\btask}[1]{\textsc{#1}}
\newcommand{\task}[1]{{\scriptsize\textsc{#1}}}
\newcommand{\tinytask}[1]{{\tiny\textsc{#1}}}
\newcommand{\vtxt}[1]{{\scriptsize\texttt{#1}}}
\newcommand{\multidec}{{\textsc{\textbf{Multi-Dec}}}\xspace}
\newcommand{\tedec}{{\textsc{\textbf{TE$\oplus$Dec}}}\xspace}
\newcommand{\teenc}{{\textsc{\textbf{TE$\oplus$Enc}}}\xspace}
\definecolor{olive}{rgb}{0.0, 0.5, 0.0}
\usepackage[pagebackref=true,breaklinks=true,colorlinks,bookmarks=false]{hyperref}

\title{Multi-Task Learning for Sequence Tagging: An Empirical Study}
\author{Soravit Changpinyo, Hexiang Hu, \and Fei Sha \\
  Department of Computer Science \\
  University of Southern California  \\
  Los Angeles, CA 90089 \\
  {\tt schangpi,hexiangh,feisha@usc.edu}}

\date{}

\begin{document}
\maketitle
\begin{abstract}

We study three general multi-task learning (MTL) approaches on 11 sequence tagging tasks. Our extensive empirical results show that in about 50\% of the cases, jointly learning all 11 tasks improves upon either independent or pairwise learning of the tasks. We also show that pairwise MTL can inform us what tasks can benefit others or what tasks can be benefited if they are learned jointly. In particular, we identify tasks that can always benefit others as well as tasks that can always be harmed by others. Interestingly, one of our MTL approaches yields embeddings of the tasks that reveal the natural clustering of semantic and syntactic tasks. Our inquiries have opened the doors to further utilization of MTL in NLP.
\end{abstract}


\section{Introduction}
\label{sIntro}

%
%
\blfootnote{
    %
    %
    \hspace{-0.65cm}  
    %
    %
    %
    %
    \hspace{-0.65cm}  
    This work is licensed under a Creative Commons 
    Attribution 4.0 International License.
    License details:
    \url{http://creativecommons.org/licenses/by/4.0/}
}

Multi-task learning (MTL) has long been studied in the machine learning literature, cf. \cite{Caruana97}. The technique has also been popular in NLP, for example, in \cite{CollobertW08,CollobertWBKKK11,LuongLSVK16}. The main thesis underpinning MTL is that solving many tasks together provides a shared inductive bias that leads to more robust and generalizable systems. This is especially appealing for NLP as data for many tasks are scarce --- shared learning thus reduces the amount of training data needed. MTL has been validated in recent work, mostly where auxiliary tasks are used to improve the performance on a target task, for example, in sequence tagging \cite{SogaardG16,BjervaPB16,PlankSG16,AlonsoP17,BingelS17}.

Despite those successful applications, several key issues about the effectiveness of MTL remain open. Firstly, with only a few exceptions, much existing work focuses on ``pairwise'' MTL where there is a target task and one or several (carefully) selected auxiliary tasks. However, \emph{can jointly learning many tasks benefit all of them together}? A positive answer will significantly raise the utility of MTL. Secondly, \emph{how are tasks related such that one could benefit another?} For instance, one plausible intuition is that syntactic and semantic tasks might benefit among their two separate groups though cross-group assistance is weak or unlikely. However,  such notions have not been put to test thoroughly on a significant number of tasks.

In this paper, we address such questions. We investigate learning jointly multiple sequence tagging tasks. Besides using independent single-task learning as a baseline and a popular shared-encoder MTL framework for sequence tagging~\cite{CollobertWBKKK11}, we propose two variants of MTL, where both the encoder and the decoder could be shared by all tasks.

We conduct extensive empirical studies on 11 sequence tagging tasks --- we defer the discussion on why we select such tasks to a later section. We demonstrate that there is a benefit to moving beyond ``pairwise" MTL. We also obtain interesting pairwise relationships that reveal which tasks are beneficial or harmful to others, and which tasks are likely to be benefited or harmed. We find such information correlated with the results of MTL using more than two tasks. We also study selecting only benefiting tasks for joint training, showing that such a ``greedy'' approach in general improves the MTL performance, highlighting the need of identifying with whom to jointly learn.
 
The rest of the paper is organized as follows. We describe different approaches for learning from multiple tasks in Sect.~\ref{sApproach}. We describe our experimental setup and results in Sect.~\ref{sSetup} and Sect.~\ref{sExpRes}, respectively. We discuss related work in Sect.~\ref{sRelated}. Finally, we conclude with discussion and future work in Sect.~\ref{sDis}.

\section{Multi-Task Learning for Sequence Tagging}
\label{sApproach}

In this section, we describe general approaches to multi-task learning (MTL) for sequence tagging. We select sequence tagging tasks for several reasons. Firstly, we want to concentrate on comparing the tasks themselves without being confounded by designing specialized MTL methods for solving complicated tasks. Sequence tagging tasks are done at the word level, allowing us to focus on simpler models while still enabling varying degrees of sharing among tasks. Secondly, those tasks are often the first steps in NLP pipelines that come with extremely diverse resources. Understanding the nature of the relationships between them is likely to have a broad impact on many downstream applications.

Let $\cst{T}$ be the number of tasks and $\mD^t$ be training data of task $t \in \{1, \ldots, \cst{T}\}$.
A dataset for each task consists of input-output pairs. In sequence tagging, each pair corresponds to a sequence of words $w_{1:\cst{L}}$ and their corresponding ground-truth tags $y_{1:\cst{L}}$, where $\cst{L}$ is the sequence length. We note that our definition of ``task" is not the same as ``domain" or ``dataset." In particular, we differentiate between tasks based on whether or not they share the label space of tags. For instance, part-of-speech tagging on weblog and that on email domains are considered the same task in this paper.

Given the training data $\{\mD^1, \ldots, \mD^\cst{T}\}$, we describe how to learn one or more models to perform all the $\cst{T}$ tasks. In general, our models follow the design of state-of-the-art sequence taggers~\cite{ReimersG17}.
They have an encoder $e$ with parameters $\theta$ that encodes a sequence of word tokens into a sequence of vectors and a decoder $d$ with parameters $\phi$ that decodes the sequence of vectors into a sequence of predicted tags $\hat{y}_{1:\cst{L}}$. That is, $c_{1:\cst{L}} = e(w_{1:\cst{L}}; \theta)$ and $\hat{y}_{1:\cst{L}} = d(c_{1:\cst{L}}; \phi)$. The model parameters are learned by minimizing some loss function $\mathcal{L}(\hat{y}_{1:\cst{L}}, y_{1:\cst{L}})$ over $\theta$ and $\phi$. In what follows, we will use superscripts to differentiate instances from different tasks.

\begin{figure*}[ht]
\centering
\begin{tabular}{cccc}
	\includegraphics[width=0.24\textwidth,trim={1cm 2cm 2.5cm 1.5cm},clip]{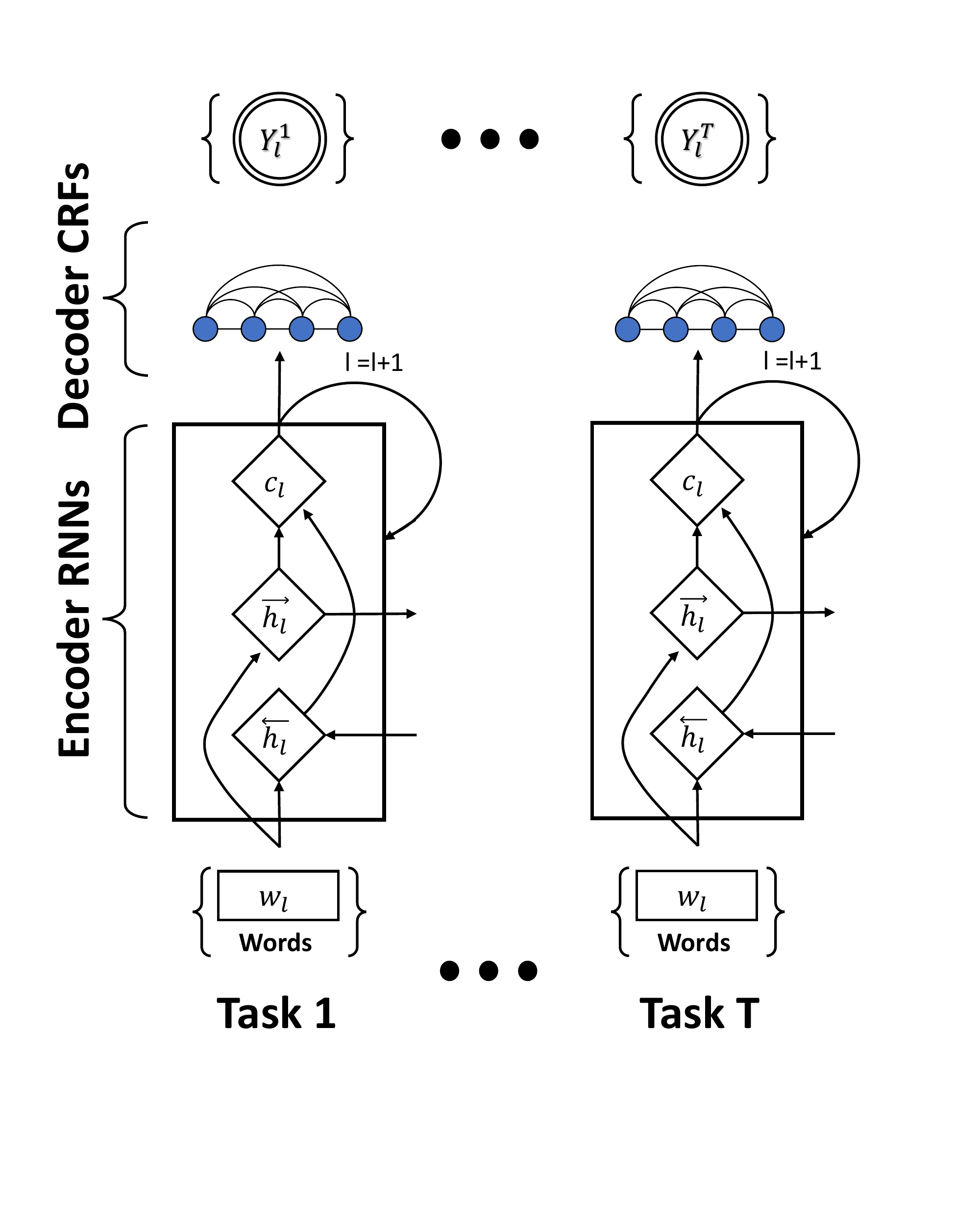} & 
	\includegraphics[width=0.24\textwidth,trim={1cm 2cm 2.5cm 1.5cm},clip]{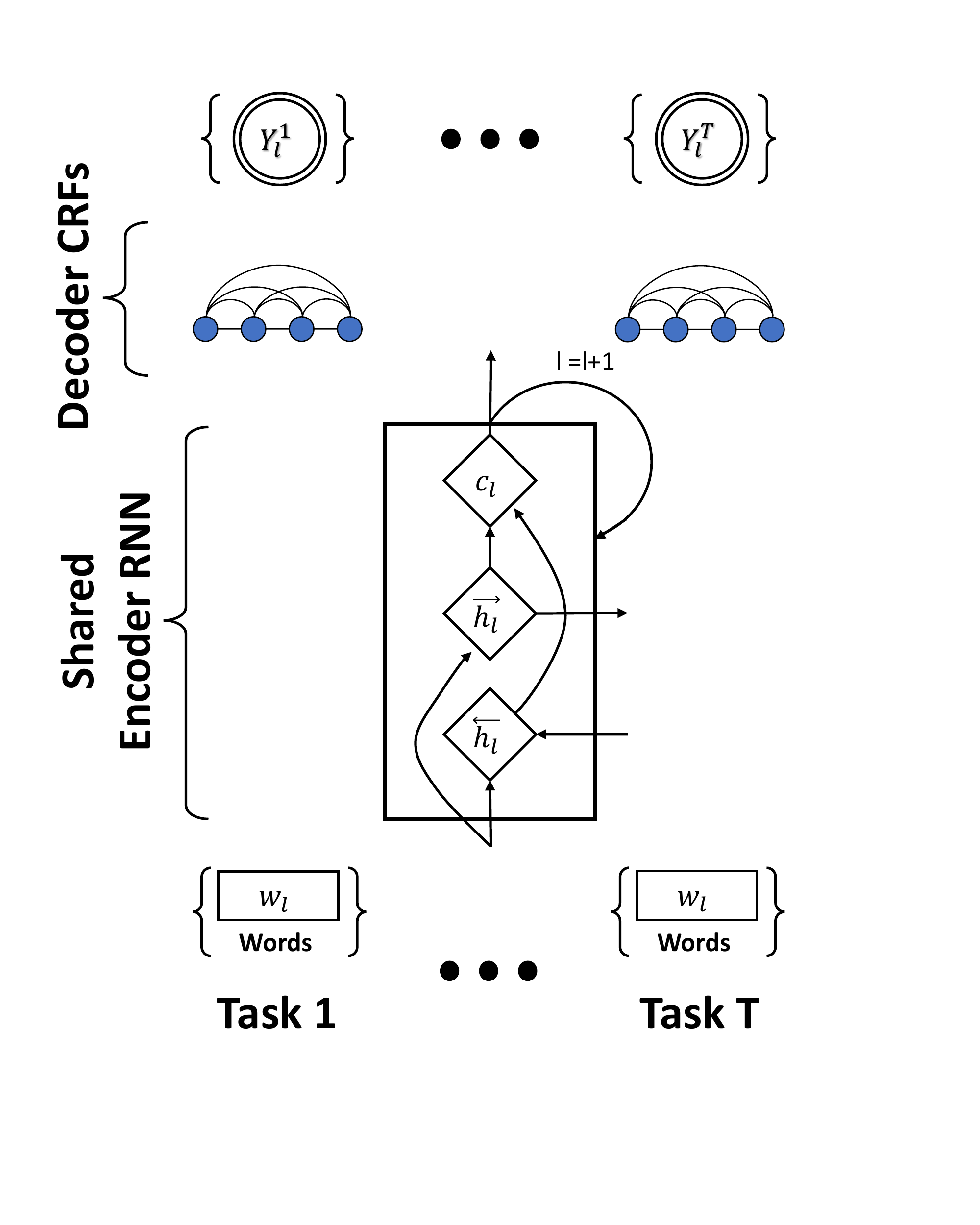}  &	
	\includegraphics[width=0.24\textwidth,trim={1cm 2cm 2.5cm 1.5cm},clip]{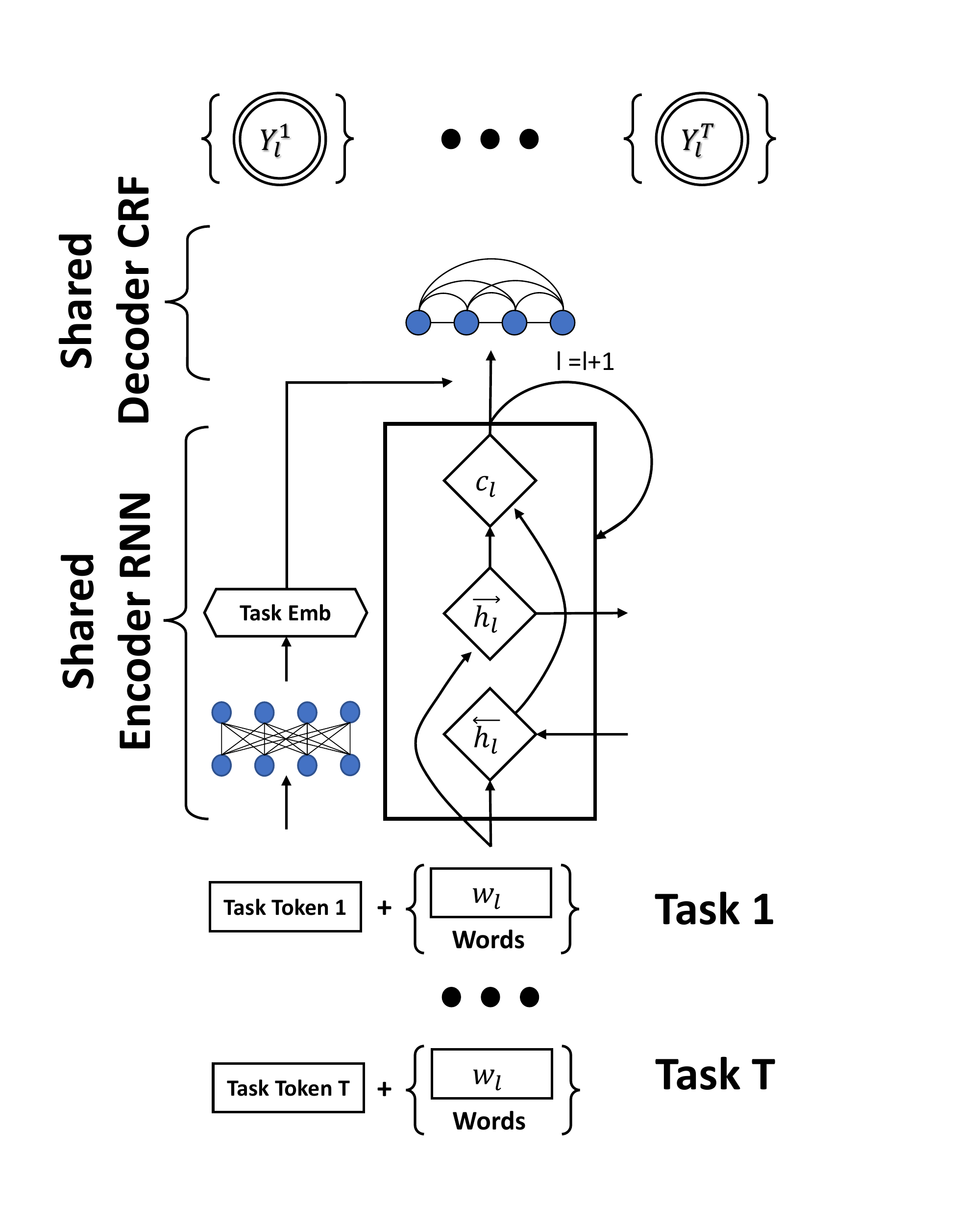} & 
	\includegraphics[width=0.24\textwidth,trim={1cm 2cm 2.5cm 1.5cm},clip]{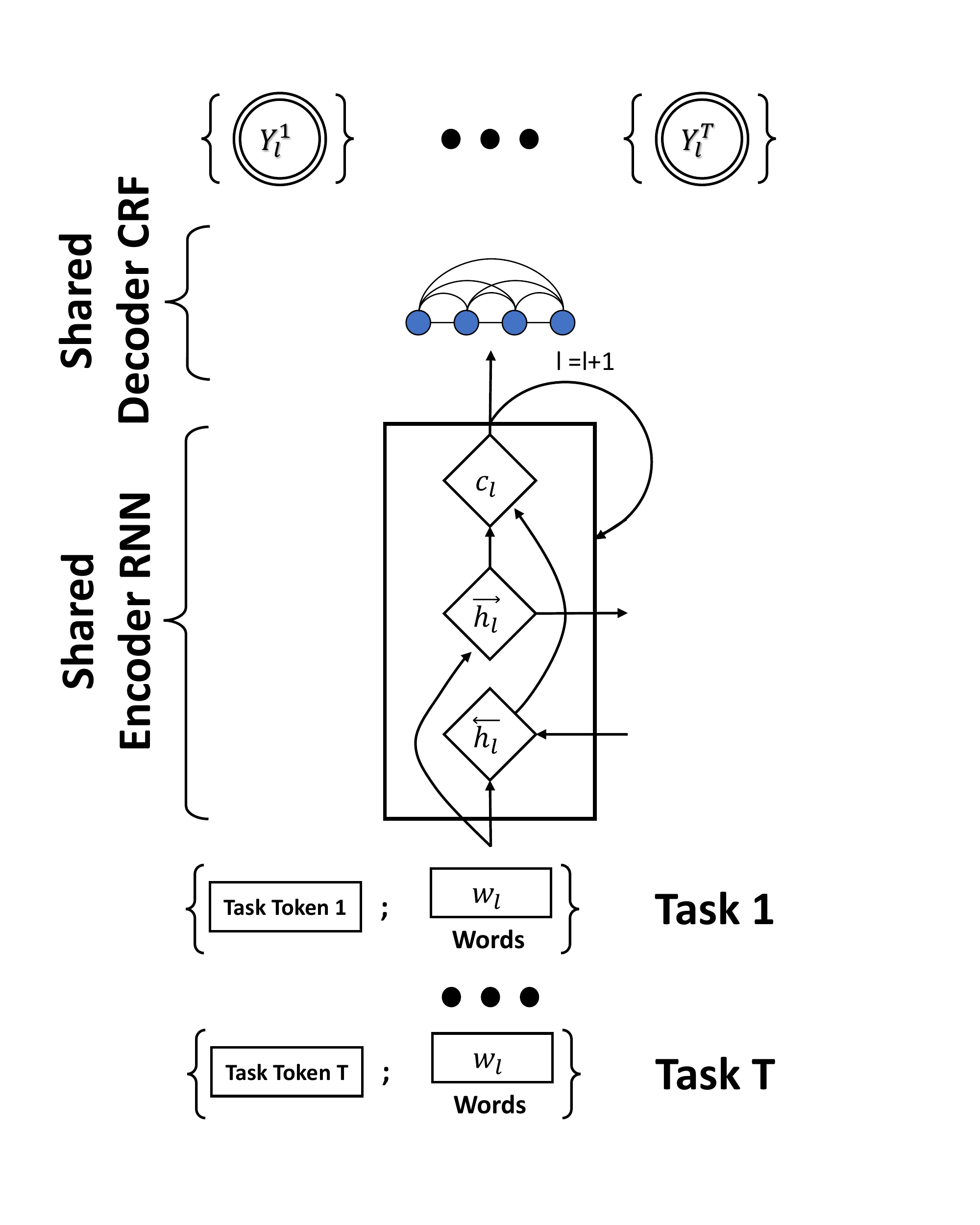} \\
	Single-task learning~\label{fSingle} & MTL (\multidec) \label{fMultiDec} &
	MTL (\tedec)~\label{fTeDec} & MTL (\teenc)~\label{fTeEnc}
\end{tabular}
\vspace{-7pt}
\caption{\small Different settings for learning from multiple tasks considered in our experiments} \label{fSettings}
\vspace{-5pt}
\end{figure*}

In \textbf{single-task learning (STL)}, we learn $\cst{T}$ models independently. For each task $t$, we have an encoder $e^t(\cdot; \theta^t)$ and a decoder $d^t(\cdot; \phi^t)$. Clearly, information is not shared between tasks in this case.

In \textbf{multi-task learning (MTL)}, we consider two or more tasks and train an MTL model \emph{jointly} over a combined loss $\sum_t\mathcal{L}(\hat{y}^t_{1:\cst{L}}, y^t_{1:\cst{L}})$. In this paper, we consider the following general frameworks that are different in the nature of how the parameters of those tasks are shared.

\noindent\textbf{Multi-task learning with multiple decoders (Multi-Dec)}
We learn a shared encoder $e(\cdot; \theta)$ and $\cst{T}$ decoders $\{d^t(\cdot; \theta^t)\}_{t=1}^{\cst{T}}$. This setting has been explored for sequence tagging in \cite{CollobertW08,CollobertWBKKK11}. In the context of sequence-to-sequence learning \cite{SutskeverVL14}, this is similar to the ``one-to-many" MTL setting in \cite{LuongLSVK16}.

\noindent\textbf{Multi-task learning with task embeddings (TE)}
We learn a shared encoder $e(\cdot; \theta)$ for the input sentence as well as a shared decoder $d(\cdot; \phi)$.
To equip our model with the ability to perform one-to-many mapping (i.e., multiple tasks), we augment the model with ``task embeddings."
Specifically, we additionally learn a function $f(t)$ that maps a task ID $t$ to a vector. We explore two ways of injecting task embeddings into models. In both cases, our $f$ is simply an embedding layer that maps the task ID to a dense vector.

One approach, denoted by \tedec, is to incorporate task embeddings into the decoder. We concatenate the task embeddings with the encoder's outputs $c^t_{1:\cst{L}}$ and then feed the result to the decoder.

The other approach, denoted by \teenc, is to combine the task embeddings  with the input sequence of words at the encoder. We implement this by prepending the task token (\texttt{<<upos>>}, \texttt{<<chunk>>}, \texttt{<<mwe>>}, etc.) to the input sequence and treat the task token as a word token~\cite{JohnsonSLKWCTVWCHD17}.

While the encoder in \tedec must learn to encode a general-purpose representation of the input sentence, the encoder in \teenc knows from the start which task it is going to perform.

Fig.~\ref{fSettings} illustrates different settings described above. Clearly, the number of model parameters is reduced significantly when we move from STL to MTL. Which MTL model is more economical depends on several factors, including the number of tasks, the dimension of the encoder output, the general architecture of the decoder, the dimension of task embeddings, how to augment the system with task embeddings, and the degree of tagset overlap.


\section{Experimental Setup}
\label{sSetup}

\subsection{Datasets and Tasks}

\begin{table*}
\centering
\scriptsize
\begin{tabular}{c|c|c|c|c|c}
Dataset & \# sentences & Token/type & Task & \# labels & Label entropy  \\ \hline
\multirow{2}{*}{Universal Dependencies v1.4} &\multirow{2}{*}{12543/16622} & \multirow{2}{*}{12.3/13.2} & \task{upos} & 17 & 2.5\\
& & & \task{xpos} & 50 & 3.1\\ \hline 
CoNLL-2000 & 8936/10948 & 12.3/13.3 & \task{chunk} & 42 & 2.3\\ \hline 
CoNLL-2003 & 14041/20744 & 9.7/11.2 & \task{ner} & 17 & 0.9\\ \hline
\multirow{2}{*}{Streusle 4.0} &\multirow{2}{*}{2723/3812} & \multirow{2}{*}{8.6/9.3} & \task{mwe} & 3 & 0.5\\
& & & \task{supsense} & 212 & 2.2\\ \hline
\multirow{2}{*}{SemCor} &\multirow{2}{*}{13851/20092} & \multirow{2}{*}{13.2/16.2} & \task{sem} & 75 & 2.2\\
& & & \task{semtr} & 11 & 1.3\\ \hline 
Broadcast News 1 & 880/1370 & 5.2/6.1 & \task{com} & 2 & 0.6\\ \hline 
FrameNet 1.5 & 3711/5711 & 8.6/9.1 & \task{frame} & 2 & 0.5\\ \hline 
Hyper-Text Corpus & 2000/3974 & 6.7/9.0 & \task{hyp} & 2 & 0.4\\ \hline  
\end{tabular}
\vspace{-7pt}
\caption{\small Datasets used in our experiments, as well as their key characteristics and their corresponding tasks. / is used to separate statistics for training data only and those for all subsets of data.} \label{tDatasets}
\vspace{-5pt}
\end{table*}

Table~\ref{tDatasets} summarizes the datasets used in our experiments, along with their corresponding tasks and important statistics. Table~\ref{tTasks} shows an example of each task's input-output pairs. We describe details below. For all tasks, we use the standard splits unless specified otherwise.

We perform universal and English-specific POS tagging (\btask{upos} and \btask{xpos}) on sentences from the English Web Treebank~\cite{BiesMWK12}, annotated by the Universal Dependencies project \cite{POS}\eat{\footnote{\url{http://universaldependencies.org/}}}. We perform syntactic chunking (\btask{chunk}) on sentences from the WSJ portion of the Penn Treebank \cite{MarcusMS93}, annotated by the CoNLL-2000 shared task \cite{Chunk}\eat{\footnote{\url{https://www.clips.uantwerpen.be/conll2000/chunking/}}}. We use sections 15-18 for training. The shared task uses section 20 for testing and does not designate the development set, so we use the first 1001 sentences for development and the rest 1011 for testing. We perform named entity recognition (\btask{ner}) on sentences from the Reuters Corpus~\cite{LewisYRL04}, consisting of news stories between August 1996-97, annotated by the CoNLL-2003 shared task \cite{NER}\eat{\footnote{\url{https://www.clips.uantwerpen.be/conll2003/ner/}}}. For both \btask{chunk} and \btask{ner}, we use the IOBES tagging scheme. 

We perform multi-word expression identification (\btask{mwe}) and supersense tagging (\btask{supsense}) on sentences from the reviews section of the English Web Treebank, annotated under the Streusle project \cite{streusle}\footnote{\url{https://github.com/nert-gu/streusle}}. We perform supersense (\btask{sem}) and semantic trait (\btask{semtr}) tagging on SemCor's sentences \cite{SemCor}, taken from a subset of the Brown Corpus \cite{FrancisK82}, using the splits provided by \cite{AlonsoP17} for both tasks\footnote{\url{https://github.com/bplank/multitasksemantics}}. For \btask{sem}, they are annotated with supersense tags \cite{MillerLTB93} by \cite{CiaramitaA06}\footnote{We consider \btask{supsense} and \btask{sem} as different tasks as they use different sets of supersense tags.}. For \btask{semtr}, \cite{AlonsoP17} uses the EuroWordNet list of ontological types for senses \cite{Vossen98} to convert supersenses into coarser semantic traits.

For sentence compression (\btask{com}), we identify which words to keep in a compressed version of sentences from the 1996 English Broadcast News Speech (HUB4) \cite{Graff97}, created by \cite{Broadcast}\footnote{\url{http://jamesclarke.net/research/resources/}}. We use the labels from the first annotator. For frame target identification (\btask{frame}), we detect words that evoke frames \cite{Das14} on sentences from the British National Corpus, annotated under the FrameNet project \cite{BakerFL98}\eat{\footnote{\url{https://framenet.icsi.berkeley.edu/fndrupal/framenet_data}}}. For both \btask{com} and \btask{frame}, we use the splits provided by \cite{BingelS17}.
For hyper-link detection (\btask{hyp}), we identify which words in the sequence are marked with hyperlinks on text from Daniel Pipes' news-style blog collected by \cite{SpitkovskyJA10}\footnote{\url{https://nlp.stanford.edu/valentin/pubs/markup-data.tar.bz2}}. We use the ``select" subset that correspond to marked, complete sentences.

\begin{table*}
\centering
\scriptsize
\begin{tabular}{c|c}
Task & Input/Output\\ \hline
\multirow{2}{*}{\task{upos}} & \vtxt{once again , thank you all for an outstanding accomplishment .} \\ 
& \vtxt{ADV ADV PUNCT VERB PRON DET ADP DET ADJ NOUN PUNCT}  \\ \hline
\multirow{2}{*}{\task{xpos}} & \vtxt{once again , thank you all for an outstanding accomplishment .} \\
& \vtxt{RB RB , VBP PRP DT IN DT JJ NN .}\\ \hline
\multirow{2}{*}{\task{chunk}} & \vtxt{the carrier also seemed eager to place blame on its american counterparts .} \\
& \vtxt{B-NP E-NP S-ADVP S-VP S-ADJP B-VP E-VP S-NP S-PP B-NP I-NP E-NP O}\\ \hline
\multirow{2}{*}{\task{ner}} & \vtxt{6. pier francesco chili ( italy ) ducati 17541}\\
& \vtxt{O B-PER I-PER E-PER O S-LOC O S-ORG O} \\ \hline
\multirow{2}{*}{\task{mwe}} & \vtxt{had to keep in mind that the a / c broke , i feel bad it was their opening !}\\
& \vtxt{B I B I I O O B I I O O O O O O O O O O}\\ \hline
\multirow{2}{*}{\task{supsense}} & \vtxt{this place may have been something sometime ; but it way past it " sell by date " .} \\
 & \vtxt{O n.GROUP O O v.stative O O O O O O p.Time p.Gestalt O v.possession p.Time n.TIME O O} \\ \hline
\multirow{2}{*}{\task{sem}} & \vtxt{a hypothetical example will illustrate this point .}\\
 & \vtxt{O adj.all noun.cognition O verb.communication O noun.communication O} \\ \hline
\multirow{2}{*}{\task{semtr}} & \vtxt{he wondered if the audience would let him finish .}\\
& \vtxt{O Mental O O Object O Agentive O BoundedEvent O}\\ \hline
\multirow{2}{*}{\task{com}} & \vtxt{he made the decisions in 1995 , in early 1996 , to spend at a very high rate .} \\
& \vtxt{KEEP KEEP DEL KEEP DEL DEL DEL DEL DEL DEL DEL KEEP KEEP KEEP KEEP DEL KEEP KEEP KEEP} \\ \hline
\multirow{2}{*}{\task{frame}} & \vtxt{please continue our important partnership .} \\
& \vtxt{O B-TARGET O B-TARGET O O} \\ \hline
\multirow{2}{*}{\task{hyp}} & \vtxt{will this incident lead to a further separation of civilizations ?} \\
& \vtxt{O O O O O O O B-HTML B-HTML B-HTML O} \\ \hline
\end{tabular}
\vspace{-7pt}
\caption{\small Examples of input-output pairs of the tasks in consideration} \label{tTasks}
\vspace{-5pt}
\end{table*}

\subsection{Metrics and Score Comparison}
\label{sMetric}
We use the span-based micro-averaged F1 score (without the O tag) for all tasks. We run each configuration three times with different initializations and compute mean and standard deviation of the scores.
To compare two scores, we use the following strategy.
Let $\mu_1$, $\sigma_1$ and $\mu_2$, $\sigma_2$ be two sets of scores (mean and std, respectively).
We say that $\mu_1$ is ``higher" than $\mu_2$ if $\mu_1 - k \times \sigma_1 > \mu_2 + k \times \sigma_2$, where $k$ is a parameter that controls how strict we want the definition to be. ``lower" is defined in the same manner with $>$ changed to $<$ and $-$ switched with $+$. $k$ is set to 1.5 in all of our experiments.

\subsection{Models}

\paragraph{General architectures} We use bidirectional recurrent neural networks (biRNNs) as our encoders for both words and characters~\cite{IrsoyC14,HuangXY15,LampleBSKD16,MaH16}. Our word/character sequence encoders and decoder classifiers are common in literature and most similar to \cite{LampleBSKD16}, but we use two-layer biRNNs (instead of one) with Gated Recurrent Unit (GRU) \cite{ChoMGBBSB14} (instead of with LSTM \cite{HochreiterS97}).

Each word is represented by a 100-dimensional vector that is the concatenation of a 50-dimensional \emph{embedding} vector and the 50-dimensional output of a \emph{character} biRNN (whose hidden representation dimension is 25 in each direction). We feed a sequence of those 100-dimensional representations to a \emph{word} biRNN, whose hidden representation dimension is 300 in each direction, resulting in a sequence of 600-dimensional vectors.  
In \tedec, the token encoder is also used to encode a task token (which is then concatenated to the encoder's output), where each task is represented as a 25-dimensional vector. For decoder/classifiers, we predict a sequence of tags using a linear projection layer (to the tagset size) followed by a conditional random field (CRF) \cite{LaffertyMP01}.

\paragraph{Implementation and training details}

We implement our models in PyTorch \cite{PyTorch} on top of the AllenNLP library \cite{AllenNLP}. Code is to be available at \url{https://github.com/schangpi/}. 

Words are lower-cased, but characters are not. 
Word embeddings are initialized with GloVe \cite{PenningtonSM14} trained on Wikipedia 2014 and Gigaword 5. We use strategies suggested by \cite{MaH16} for initializing other parameters in our networks. Character embeddings are initialized uniformly in $[-\sqrt{3/d}, \sqrt{3/d}]$, where $d$ is the dimension of the embeddings. Weight matrices are initialized with Xavier Uniform \cite{GlorotB10}, i.e., uniformly in $[-\sqrt{6/(r+c)}, \sqrt{6/(r+c)}]$, where $r$ and $c$ are the number of of rows and columns in the structure. Bias vectors are initialized with zeros.

We use Adam \cite{Adam} with default hyperparameters and a mini-batch size of 32. The dropout rate is 0.25 for the character encoder and 0.5 for the word encoder. We use gradient normalization \cite{PascanuMB13} with a threshold of 5. We halve the learning rate  if the validation performance does not improve for two epochs, and stop training if the validation performance does not improve for 10 epochs. We use L2 regularization with parameter 0.01 for the transition matrix of the CRF.

For the training of MTL models, we make sure that each mini-batch is balanced; the difference in numbers of examples from any pair of tasks is no more than 1. As a result, each epoch may not go through all examples of some tasks whose training set sizes are large. In a similar manner, during validation, the average F1 score is over all tasks rather than over all validation examples. 

\subsection{Various Settings for Learning from Multiple Tasks} 
\label{sTrainSettings}
We consider the following settings: (i) ``STL" where we train each model on one task alone; (ii) ``Pairwise MTL" where we train on two tasks jointly; (iii) ``All MTL" where we train on all tasks jointly; (iv) ``Oracle MTL" where we train on the Oracle set of the testing task jointly with the testing task; (v) ``All-but-one MTL" setting where we train on all tasks jointly except for one (as part of Sect.~\ref{sAnalysis}.)

\paragraph{Constructing the Oracle Set of a Testing Task} 
The Oracle set of a task $t$ is constructed from the pairwise performances:  let $\mu(A, t), \sigma(A, t)$ be the F1 score and the standard deviation of a model that is jointly trained on a set of tasks in the set $A$ and that is tested on task $t$. Task $s$ is considered ``beneficial" to another (testing) task $t$ if $\mu(\{s, t\}, t)$ is ``higher" than $\mu(\{t\}, t)$ (cf. Sect.~\ref{sMetric}). Then, the ``Oracle" set for a task $t$ is the set of its all beneficial (single) tasks. Throughout our experiments, we compute $\mu$ and $\sigma$ by averaging over three rounds (cf. Sect.~\ref{sMetric}, standard deviations can be found in Appendix~\ref{sSuppStd}.)

\section{Results and Analysis}
\label{sExpRes}

\subsection{Main Results}

\begin{figure*}[ht]
\centering
\includegraphics[width=0.32\textwidth]{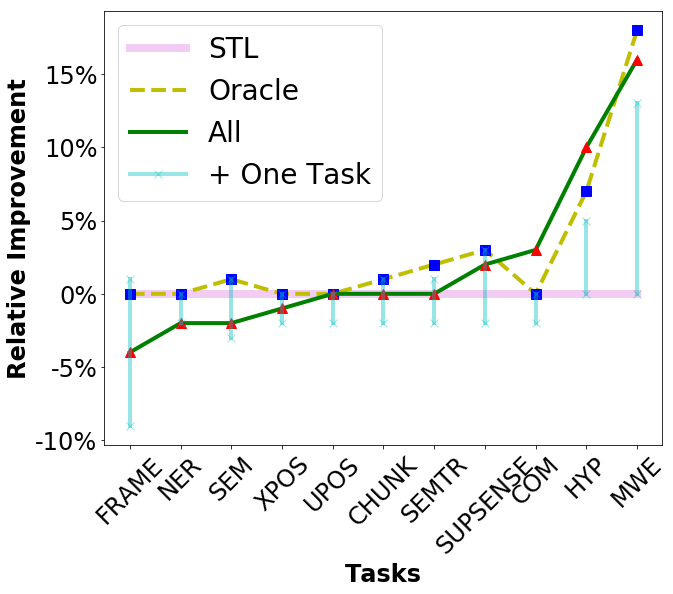}
\includegraphics[width=0.32\textwidth]{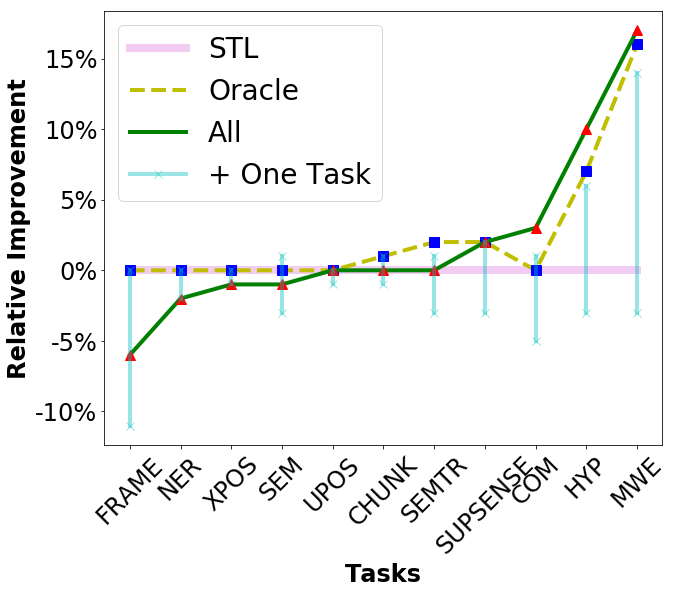}
\includegraphics[width=0.32\textwidth]{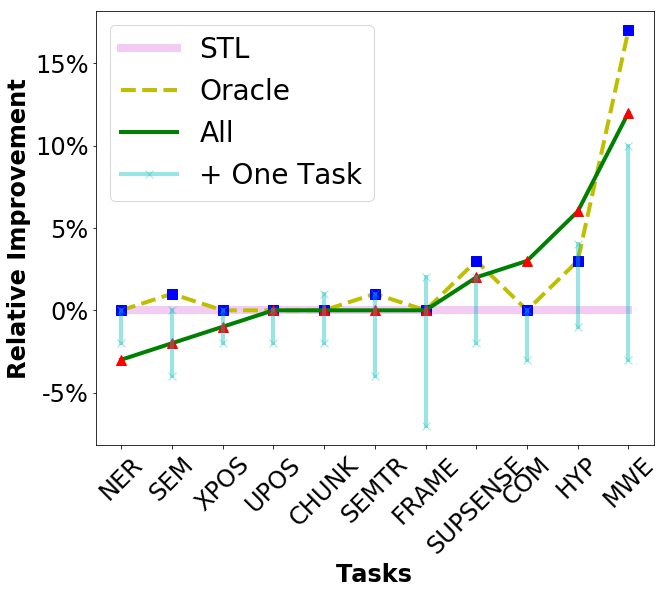}
\vspace{-7pt}
\caption{\small Summary of our results for MTL methods \multidec (left), \tedec (middle), and \teenc (right) on various settings with different types of sharing. The vertical axis is the relative improvement over STL. See texts for details. Best viewed in color.} \label{fSummary}
\vspace{-10pt}
\end{figure*}

Fig.~\ref{fSummary} summarizes our main findings. We compare relative improvement over single-task learning (STL) between various settings with different types of sharing in Sect.~\ref{sTrainSettings}. Scores from the pairwise setting (``+One Task'') are represented as a vertical bar, delineating the maximum and minimum improvement over STL by jointly learning a task with one of the remaining 10 tasks. The ``All'' setting (red triangles) indicates the joint learning all 11 tasks. The ``Oracle'' setting (blue rectangles) indicates the joint learning using a subset of 11 tasks which are deemed beneficial, based on corresponding performances in pairwise MTL, as defined in Sect.~\ref{sTrainSettings}.

We observe that (1) [STL vs. Pairwise/All] Neither pairwise MTL nor All always improves upon STL; (2) [STL vs. Oracle] Oracle in general outperforms or at least does not worsen STL;
(3) [All/Oracle vs. Pairwise] All does better than Pairwise on about half of the cases, while Oracle almost always does better than Pairwise;
(4) [All vs. Oracle] Consider when both All and Oracle improve upon STL. For \multidec and \teenc, Oracle generally dominates All, except on the task \btask{hyp}. For \tedec, their magnitudes of improvement are mostly comparable, except on \btask{semtr} (Oracle is better) and on \btask{hyp} (All is better). In addition, All is better than Oracle on the task \btask{com}, in which case Oracle is STL. 

In Appendix~\ref{sSuppCompareApp}, we compare different MTL approaches: \multidec, \tedec, and \teenc. There is no significant difference among them.

\subsection{Pairwise MTL results}
\label{sExpPwRes}

\begin{figure*}[t]
\centering
\includegraphics[width=0.32\textwidth]{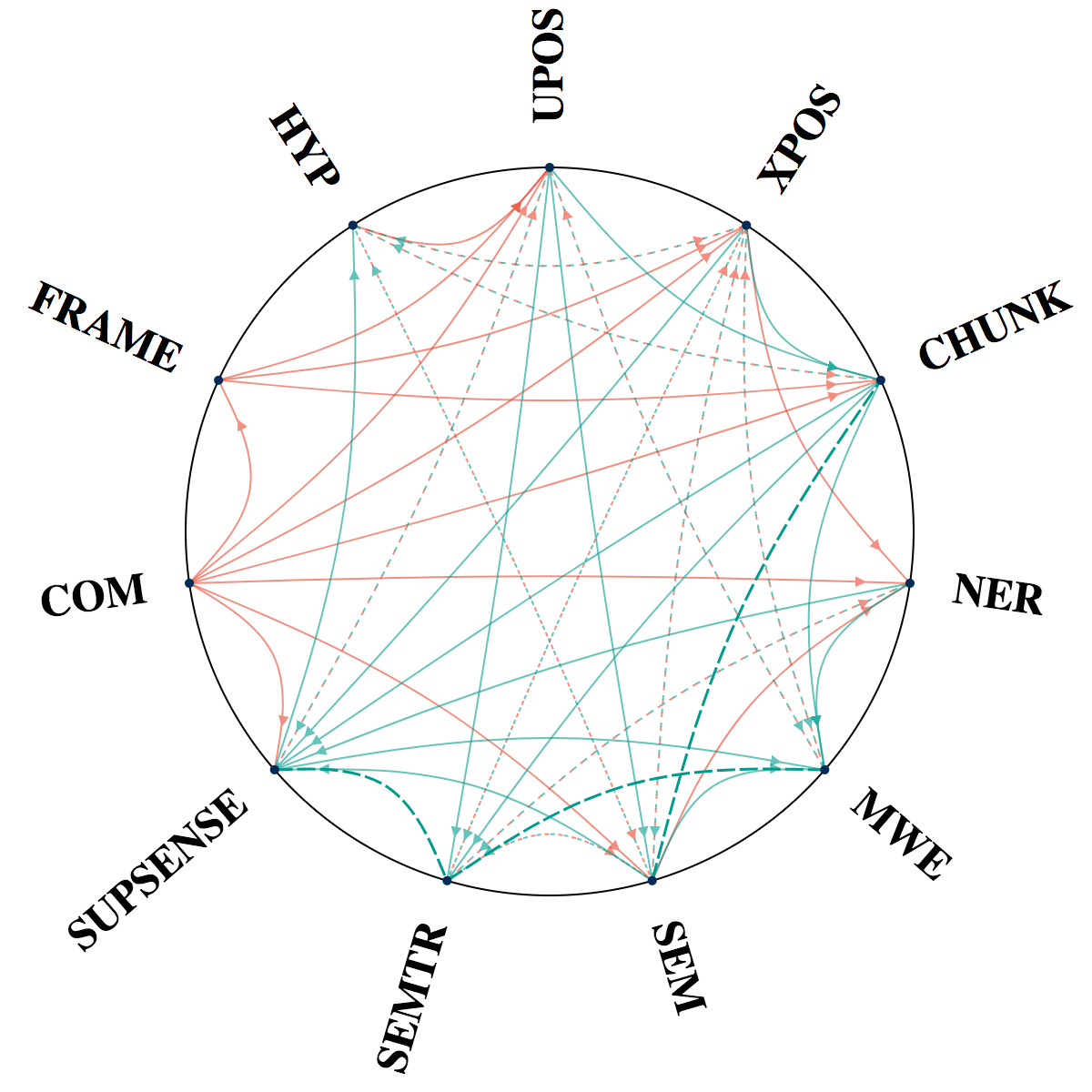}
\includegraphics[width=0.32\textwidth]{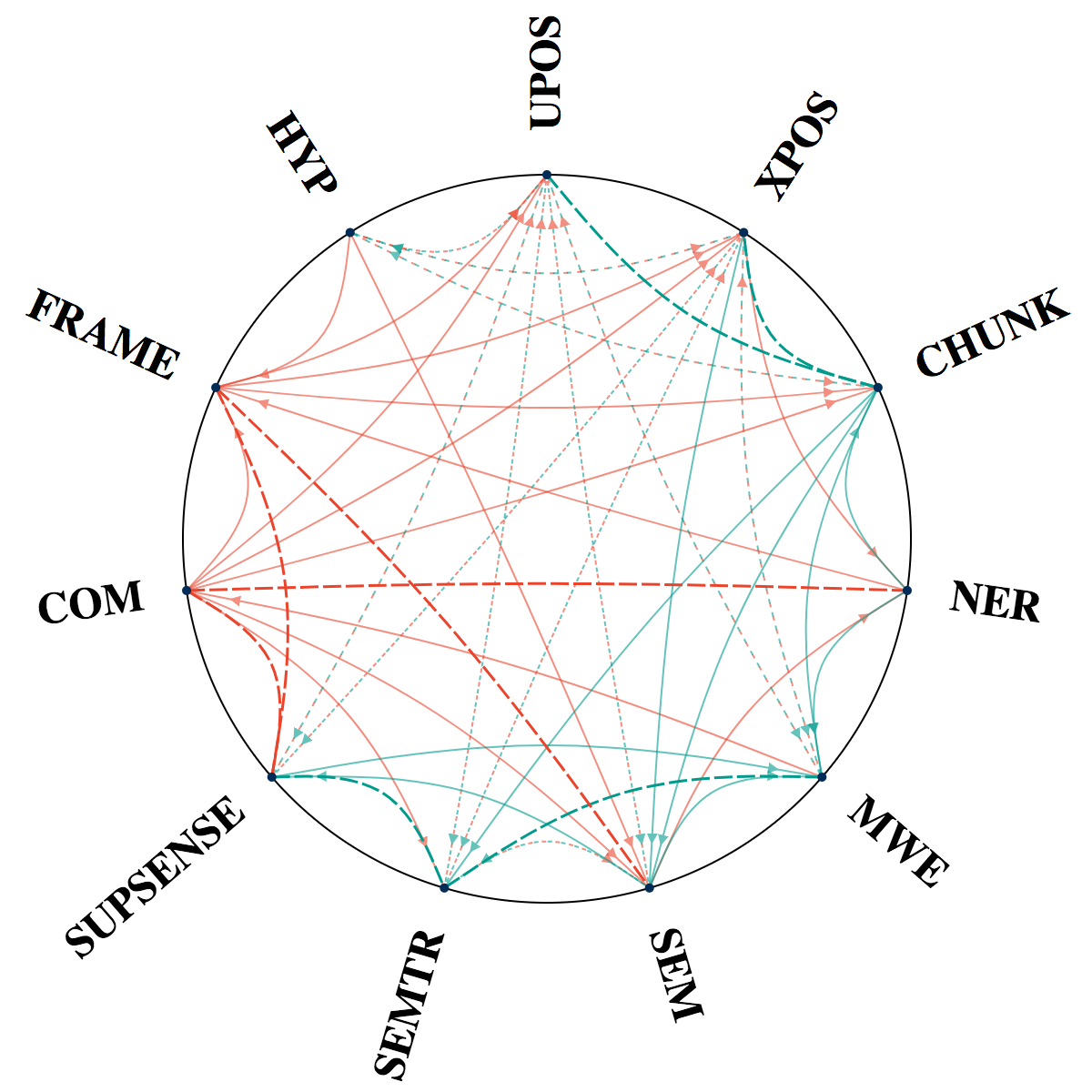}
\includegraphics[width=0.32\textwidth]{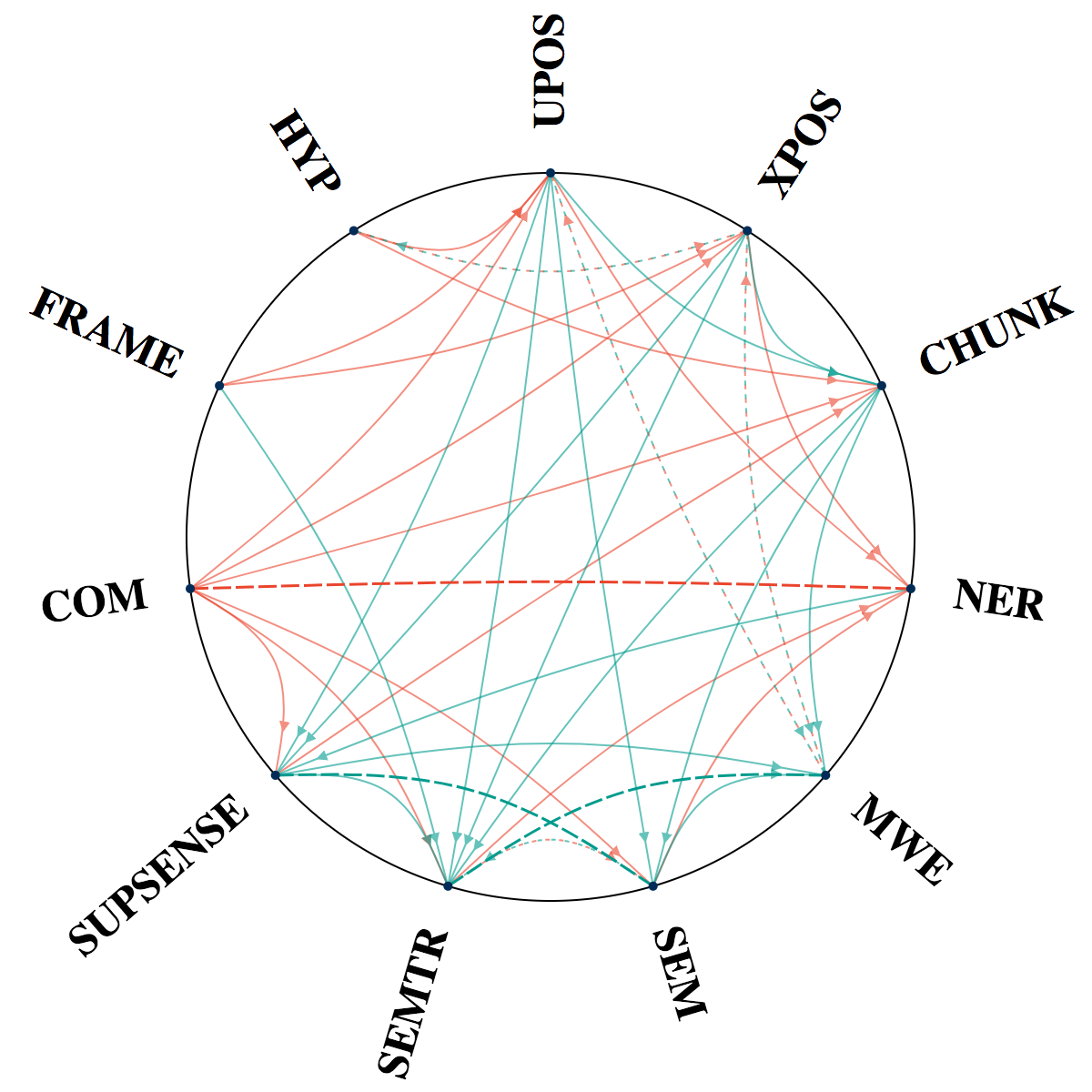}
\vspace{-7pt}
\caption{\small Pairwise MTL relationships (benefit vs. harm) using \multidec (left), \tedec (middle), and \teenc (right). \textbf{Solid} green (red) directed edge from $s$ to $t$ denotes $s$ benefiting (harming) $t$. \textbf{Dashed} Green (Red) edges between $s$ and $t$ denote they benefiting (harming) \emph{each other}. \textbf{Dotted} edges denote asymmetric relationship: benefit in one direction but harm in the reverse direction. Absence of edges denotes neutral relationships. \textit{Best viewed in color and with a zoom-in.}} \label{fTaskRel}
\vspace{-5pt}
\end{figure*}

\paragraph{Summary} The summary plot in Fig.~\ref{fTaskRel} gives a bird's-eye view of patterns in which a task might benefit or harm another one. For example, \textsc{mwe} is always benefited from jointly learning any of the 10 tasks as the \emph{incoming edges} are green, so is \textsc{semtr} in most cases. On the other end, \textsc{com} seems to be harming any of the 10 as the \emph{outgoing edges} are almost always red. For \textsc{chunk} and \textsc{u/xpos}, it generally benefits others (or at least does not do harm) as most of their \emph{outgoing} edges are green.  

In Table~\ref{tPwMultiDec}-\ref{tPwTEEnc}, we report F1 scores for \multidec, \tedec, and \teenc, respectively. In each table, rows denote settings in which we train our models, and columns correspond to tasks we test them on. We also include ``Average" of all pairwise scores, as well as the number of positive ({\color{olive}$\uparrow$}) and negative ({\color{red}$\downarrow$}) relationships in each row or each column.

\paragraph{Which tasks are benefited or harmed by others in pairwise MTL?}
\btask{mwe}, \btask{supsense}, \btask{semtr}, and \btask{hyp} are generally benefited by other tasks. The improvement is more significant in \btask{mwe} and \btask{hyp}.
\btask{upos}, \btask{xpos}, \btask{ner}, \btask{com}, and \btask{frame} (\multidec and \tedec) are often hurt by other tasks.
Finally, the results are mixed for \btask{chunk} and \btask{sem}.

\paragraph{Which tasks are beneficial or harmful?}
\btask{upos}, \btask{xpos}, and \btask{chunk} are universal helpers, beneficial in 16, 17, and 14 cases, while harmful only in 1, 3, and 0 cases, respectively. Interestingly, \btask{chunk} never hurts any task, while both \btask{upos} and \btask{xpos} can be harmful to \btask{ner}. While \btask{chunk} is considered more of a syntactic task, the fact that it informs other tasks about the boundaries of phrases may aid the learning of other semantic tasks (task embeddings in Sect.~\ref{sAnalysis} seem to support this hypothesis).

On the other hand, \btask{com}, \btask{frame}, and \btask{hyp} are generally harmful, all useful in 0 cases and causing the performance drop in 22, 10, 12 cases, respectively. One factor that may play a role is the training set sizes of these tasks. However, we note that both \btask{mwe} and \btask{supsense} (Streusle dataset) has smaller training set sizes than \btask{frame} does, but those tasks can still benefit some tasks. (On the other hand, \btask{ner} has the largest training set, but infrequently benefits other tasks, less frequently than \btask{supsense} does.) Another potential cause is the fact that all those harmful tasks have the smallest label size of 2. This combined with small dataset sizes leads to a higher chance of overfitting. Finally, it may be possible that harmful tasks are simply unrelated; for example, the nature of \btask{com}, \btask{frame}, or \btask{hyp} may be very different from other tasks --- an entirely different kind of reasoning is required.   

Finally, \btask{ner}, \btask{mwe}, \btask{sem}, \btask{semtr}, and \btask{supsense} can be beneficial or harmful, depending on which other tasks they are trained with.

\begin{table*}[ht]
\centering
\scriptsize{
\begin{tabular}{c|c|c|c|c|c|c|c|c|c|c|c|c|c}
 & \tinytask{upos} & \tinytask{xpos} & \tinytask{chunk} & \tinytask{ner} & \tinytask{mwe} & \tinytask{sem} & \tinytask{semtr} & \tinytask{supsense} & \tinytask{com} & \tinytask{frame} & \tinytask{hyp} & {\color{olive}\tiny{\#$\uparrow$}}  & {\color{red}\tiny{\#$\downarrow$}} \\ \hline
+\tinytask{upos} & {\color{blue}\scriptsize{95.4}}& \tiny{95.01}& {\color{olive}\scriptsize{94.18} $\uparrow$ }& \tiny{87.68}& {\color{olive}\scriptsize{59.99} $\uparrow$ }& {\color{olive}\scriptsize{73.23} $\uparrow$ }& {\color{olive}\scriptsize{74.93} $\uparrow$ }& {\color{olive}\scriptsize{68.25} $\uparrow$ }& \tiny{72.46}& \tiny{62.14}& \tiny{48.02}& {\tiny{5}} & {\tiny{0}}\\ \hline
+\tinytask{xpos} & \tiny{95.38}& {\color{blue}\scriptsize{95.04}}& {\color{olive}\scriptsize{93.97} $\uparrow$ }& {\color{red}\scriptsize{87.61} $\downarrow$ }& {\color{olive}\scriptsize{58.87} $\uparrow$ }& {\color{olive}\scriptsize{73.34} $\uparrow$ }& {\color{olive}\scriptsize{74.91} $\uparrow$ }& {\color{olive}\scriptsize{67.78} $\uparrow$ }& \tiny{72.83}& \tiny{60.77}& {\color{olive}\scriptsize{48.81} $\uparrow$ }& {\tiny{6}} & {\tiny{1}}\\ \hline
+\tinytask{chunk} & \tiny{95.43}& \tiny{95.1}& {\color{blue}\scriptsize{93.49}}& \tiny{87.96}& {\color{olive}\scriptsize{59.18} $\uparrow$ }& {\color{olive}\scriptsize{73.16} $\uparrow$ }& {\color{olive}\scriptsize{74.79} $\uparrow$ }& {\color{olive}\scriptsize{67.39} $\uparrow$ }& \tiny{72.44}& \tiny{62.67}& {\color{olive}\scriptsize{47.85} $\uparrow$ }& {\tiny{5}} & {\tiny{0}}\\ \hline
+\tinytask{ner} & \tiny{95.38}& \tiny{94.98}& \tiny{93.47}& {\color{blue}\scriptsize{88.24}}& {\color{olive}\scriptsize{55.4} $\uparrow$ }& \tiny{72.88}& {\color{olive}\scriptsize{74.34} $\uparrow$ }& {\color{olive}\scriptsize{68.06} $\uparrow$ }& \tiny{70.93}& \tiny{62.39}& \tiny{47.9}& {\tiny{3}} & {\tiny{0}}\\ \hline
+\tinytask{mwe} & {\color{red}\scriptsize{95.15} $\downarrow$ }& {\color{red}\scriptsize{94.7} $\downarrow$ }& \tiny{93.54}& \tiny{88.15}& {\color{blue}\scriptsize{53.07}}& \tiny{72.75}& {\color{olive}\scriptsize{74.51} $\uparrow$ }& \tiny{66.88}& \tiny{71.31}& \tiny{61.75}& \tiny{47.32}& {\tiny{1}} & {\tiny{2}}\\ \hline
+\tinytask{sem} & \tiny{95.23}& {\color{red}\scriptsize{94.77} $\downarrow$ }& {\color{olive}\scriptsize{93.63} $\uparrow$ }& {\color{red}\scriptsize{87.35} $\downarrow$ }& {\color{olive}\scriptsize{60.16} $\uparrow$ }& {\color{blue}\scriptsize{72.77}}& {\color{olive}\scriptsize{74.73} $\uparrow$ }& {\color{olive}\scriptsize{68.29} $\uparrow$ }& \tiny{72.72}& \tiny{61.74}& {\color{olive}\scriptsize{48.15} $\uparrow$ }& {\tiny{5}} & {\tiny{2}}\\ \hline
+\tinytask{semtr} & \tiny{95.17}& {\color{red}\scriptsize{94.86} $\downarrow$ }& \tiny{93.61}& {\color{red}\scriptsize{87.34} $\downarrow$ }& {\color{olive}\scriptsize{58.84} $\uparrow$ }& {\color{red}\scriptsize{72.5} $\downarrow$ }& {\color{blue}\scriptsize{74.02}}& {\color{olive}\scriptsize{68.6} $\uparrow$ }& \tiny{71.96}& \tiny{62.03}& \tiny{47.74}& {\tiny{2}} & {\tiny{3}}\\ \hline
+\tinytask{supsense} & {\color{red}\scriptsize{95.08} $\downarrow$ }& \tiny{94.75}& \tiny{93.2}& \tiny{87.9}& {\color{olive}\scriptsize{58.81} $\uparrow$ }& \tiny{72.81}& {\color{olive}\scriptsize{74.61} $\uparrow$ }& {\color{blue}\scriptsize{66.81}}& \tiny{72.24}& \tiny{61.94}& {\color{olive}\scriptsize{49.23} $\uparrow$ }& {\tiny{3}} & {\tiny{1}}\\ \hline
+\tinytask{com} & {\color{red}\scriptsize{93.04} $\downarrow$ }& {\color{red}\scriptsize{93.19} $\downarrow$ }& {\color{red}\scriptsize{91.94} $\downarrow$ }& {\color{red}\scriptsize{86.62} $\downarrow$ }& \tiny{53.89}& {\color{red}\scriptsize{70.39} $\downarrow$ }& \tiny{72.6}& {\color{red}\scriptsize{65.57} $\downarrow$ }& {\color{blue}\scriptsize{72.71}}& {\color{red}\scriptsize{56.52} $\downarrow$ }& \tiny{47.41}& {\tiny{0}} & {\tiny{7}}\\ \hline
+\tinytask{frame} & {\color{red}\scriptsize{94.98} $\downarrow$ }& {\color{red}\scriptsize{94.64} $\downarrow$ }& {\color{red}\scriptsize{93.22} $\downarrow$ }& \tiny{88.15}& \tiny{53.88}& \tiny{72.76}& \tiny{74.18}& \tiny{66.59}& \tiny{72.47}& {\color{blue}\scriptsize{62.04}}& \tiny{47.5}& {\tiny{0}} & {\tiny{3}}\\ \hline
+\tinytask{hyp} & {\color{red}\scriptsize{94.84} $\downarrow$ }& {\color{red}\scriptsize{94.46} $\downarrow$ }& {\color{red}\scriptsize{92.96} $\downarrow$ }& \tiny{87.98}& \tiny{53.08}& {\color{red}\scriptsize{72.47} $\downarrow$ }& \tiny{74.23}& \tiny{66.47}& \tiny{71.82}& \tiny{61.02}& {\color{blue}\scriptsize{46.73}}& {\tiny{0}} & {\tiny{4}}\\ \hline \hline
 {\color{olive}\tiny{\#$\uparrow$}} & {\tiny{0}}& {\tiny{0}}& {\tiny{3}}& {\tiny{0}}& {\tiny{7}}& {\tiny{3}}& {\tiny{7}}& {\tiny{6}}& {\tiny{0}}& {\tiny{0}}& {\tiny{4}}\\ \hline
 {\color{red}\tiny{\#$\downarrow$}} & {\tiny{5}}& {\tiny{6}}& {\tiny{3}}& {\tiny{4}}& {\tiny{0}}& {\tiny{3}}& {\tiny{0}}& {\tiny{1}}& {\tiny{0}}& {\tiny{1}}& {\tiny{0}}\\ \hline \hline
Average & 94.97& 94.65& 93.37& 87.67& 57.21& 72.63& 74.38& 67.39& 72.12& 61.3& 47.99& & \\ \hline 
All & {\color{red}\scriptsize{95.04} $\downarrow$ }& {\color{red}\scriptsize{94.31} $\downarrow$ }& \tiny{93.44}& {\color{red}\scriptsize{86.38} $\downarrow$ }& {\color{olive}\scriptsize{61.43} $\uparrow$ }& {\color{red}\scriptsize{71.53} $\downarrow$ }& {\color{olive}\scriptsize{74.26} $\uparrow$ }& {\color{olive}\scriptsize{68.1} $\uparrow$ }& \tiny{74.54}& \tiny{59.71}& {\color{olive}\scriptsize{51.41} $\uparrow$ }& {\tiny{4}} & {\tiny{4}}\\ \hline 
Oracle&  {\color{blue}\tiny{95.4}}&  {\color{blue}\tiny{95.04}}& {\color{olive}\scriptsize{94.01} $\uparrow$ }&  {\color{blue}\tiny{88.24}}& {\color{olive}\scriptsize{62.76} $\uparrow$ }& {\color{olive}\scriptsize{73.32} $\uparrow$ }& {\color{olive}\scriptsize{75.23} $\uparrow$ }& {\color{olive}\scriptsize{68.53} $\uparrow$ }&  {\color{blue}\tiny{72.71}}&  {\color{blue}\tiny{62.04}}& {\color{olive}\scriptsize{50.0} $\uparrow$ }& {\tiny{6}} & {\tiny{0}}\\ \hline
\end{tabular}
\vspace{-7pt}
\caption{\small F1 scores for \multidec. We compare STL setting (blue), with pairwise MTL (+$\langle$task$\rangle$), All, and Oracle. We test on each task in the columns. Beneficial settings are in green. Harmful setting are in red. The last two columns indicate how many tasks are helped or harmed by the task at that row.}\label{tPwMultiDec}}
\vspace{-5pt}
\end{table*}
\begin{table*}[ht]
\centering
\scriptsize{
\begin{tabular}{c|c|c|c|c|c|c|c|c|c|c|c|c|c}
 & \tinytask{upos} & \tinytask{xpos} & \tinytask{chunk} & \tinytask{ner} & \tinytask{mwe} & \tinytask{sem} & \tinytask{semtr} & \tinytask{supsense} & \tinytask{com} & \tinytask{frame} & \tinytask{hyp} & {\color{olive}\tiny{\#$\uparrow$}}  & {\color{red}\tiny{\#$\downarrow$}} \\ \hline
+\tinytask{upos} & {\color{blue}\scriptsize{95.4}}& \tiny{94.99}& {\color{olive}\scriptsize{94.02} $\uparrow$ }& \tiny{87.99}& {\color{olive}\scriptsize{60.28} $\uparrow$ }& {\color{olive}\scriptsize{73.17} $\uparrow$ }& {\color{olive}\scriptsize{74.87} $\uparrow$ }& {\color{olive}\scriptsize{67.8} $\uparrow$ }& \tiny{72.86}& \tiny{61.54}& {\color{olive}\scriptsize{49.36} $\uparrow$ }& {\tiny{6}} & {\tiny{0}}\\ \hline
+\tinytask{xpos} & \tiny{95.4}& {\color{blue}\scriptsize{95.04}}& {\color{olive}\scriptsize{94.18} $\uparrow$ }& {\color{red}\scriptsize{87.65} $\downarrow$ }& {\color{olive}\scriptsize{60.32} $\uparrow$ }& {\color{olive}\scriptsize{73.21} $\uparrow$ }& {\color{olive}\scriptsize{74.84} $\uparrow$ }& {\color{olive}\scriptsize{68.3} $\uparrow$ }& \tiny{72.87}& \tiny{61.44}& {\color{olive}\scriptsize{49.23} $\uparrow$ }& {\tiny{6}} & {\tiny{1}}\\ \hline
+\tinytask{chunk} & {\color{olive}\scriptsize{95.57} $\uparrow$ }& {\color{olive}\scriptsize{95.21} $\uparrow$ }& {\color{blue}\scriptsize{93.49}}& \tiny{88.11}& {\color{olive}\scriptsize{57.61} $\uparrow$ }& {\color{olive}\scriptsize{73.02} $\uparrow$ }& {\color{olive}\scriptsize{74.73} $\uparrow$ }& \tiny{67.29}& \tiny{73.3}& \tiny{61.39}& {\color{olive}\scriptsize{48.43} $\uparrow$ }& {\tiny{6}} & {\tiny{0}}\\ \hline
+\tinytask{ner} & \tiny{95.32}& \tiny{95.09}& {\color{olive}\scriptsize{93.64} $\uparrow$ }& {\color{blue}\scriptsize{88.24}}& {\color{olive}\scriptsize{55.17} $\uparrow$ }& \tiny{72.77}& \tiny{74.01}& \tiny{67.25}& {\color{red}\scriptsize{71.08} $\downarrow$ }& {\color{red}\scriptsize{59.25} $\downarrow$ }& \tiny{48.24}& {\tiny{2}} & {\tiny{2}}\\ \hline
+\tinytask{mwe} & {\color{red}\scriptsize{95.11} $\downarrow$ }& {\color{red}\scriptsize{94.8} $\downarrow$ }& \tiny{93.59}& \tiny{87.99}& {\color{blue}\scriptsize{53.07}}& \tiny{72.66}& {\color{olive}\scriptsize{74.63} $\uparrow$ }& \tiny{66.88}& {\color{red}\scriptsize{70.93} $\downarrow$ }& \tiny{56.77}& \tiny{45.83}& {\tiny{1}} & {\tiny{3}}\\ \hline
+\tinytask{sem} & {\color{red}\scriptsize{95.2} $\downarrow$ }& \tiny{94.82}& \tiny{93.45}& {\color{red}\scriptsize{87.27} $\downarrow$ }& {\color{olive}\scriptsize{58.21} $\uparrow$ }& {\color{blue}\scriptsize{72.77}}& {\color{olive}\scriptsize{74.72} $\uparrow$ }& {\color{olive}\scriptsize{68.46} $\uparrow$ }& \tiny{73.14}& {\color{red}\scriptsize{60.09} $\downarrow$ }& \tiny{47.95}& {\tiny{3}} & {\tiny{3}}\\ \hline
+\tinytask{semtr} & {\color{red}\scriptsize{95.21} $\downarrow$ }& {\color{red}\scriptsize{94.8} $\downarrow$ }& \tiny{93.47}& \tiny{87.75}& {\color{olive}\scriptsize{58.55} $\uparrow$ }& {\color{red}\scriptsize{72.5} $\downarrow$ }& {\color{blue}\scriptsize{74.02}}& {\color{olive}\scriptsize{68.18} $\uparrow$ }& \tiny{71.74}& \tiny{59.77}& \tiny{46.96}& {\tiny{2}} & {\tiny{3}}\\ \hline
+\tinytask{supsense} & {\color{red}\scriptsize{95.05} $\downarrow$ }& {\color{red}\scriptsize{94.81} $\downarrow$ }& \tiny{93.25}& \tiny{87.94}& {\color{olive}\scriptsize{58.75} $\uparrow$ }& \tiny{72.71}& {\color{olive}\scriptsize{74.52} $\uparrow$ }& {\color{blue}\scriptsize{66.81}}& {\color{red}\scriptsize{69.13} $\downarrow$ }& {\color{red}\scriptsize{55.68} $\downarrow$ }& \tiny{47.29}& {\tiny{2}} & {\tiny{4}}\\ \hline
+\tinytask{com} & {\color{red}\scriptsize{94.03} $\downarrow$ }& {\color{red}\scriptsize{93.94} $\downarrow$ }& {\color{red}\scriptsize{92.29} $\downarrow$ }& {\color{red}\scriptsize{86.59} $\downarrow$ }& \tiny{51.72}& {\color{red}\scriptsize{70.37} $\downarrow$ }& {\color{red}\scriptsize{71.76} $\downarrow$ }& {\color{red}\scriptsize{64.98} $\downarrow$ }& {\color{blue}\scriptsize{72.71}}& {\color{red}\scriptsize{55.25} $\downarrow$ }& \tiny{45.24}& {\tiny{0}} & {\tiny{8}}\\ \hline
+\tinytask{frame} & {\color{red}\scriptsize{94.79} $\downarrow$ }& {\color{red}\scriptsize{94.66} $\downarrow$ }& {\color{red}\scriptsize{93.23} $\downarrow$ }& \tiny{88.02}& \tiny{53.05}& {\color{red}\scriptsize{72.26} $\downarrow$ }& \tiny{74.21}& {\color{red}\scriptsize{66.2} $\downarrow$ }& \tiny{72.89}& {\color{blue}\scriptsize{62.04}}& \tiny{46.0}& {\tiny{0}} & {\tiny{5}}\\ \hline
+\tinytask{hyp} & {\color{red}\scriptsize{94.35} $\downarrow$ }& {\color{red}\scriptsize{94.56} $\downarrow$ }& {\color{red}\scriptsize{92.86} $\downarrow$ }& \tiny{87.91}& \tiny{52.98}& {\color{red}\scriptsize{72.15} $\downarrow$ }& \tiny{74.19}& \tiny{66.52}& \tiny{70.47}& {\color{red}\scriptsize{55.35} $\downarrow$ }& {\color{blue}\scriptsize{46.73}}& {\tiny{0}} & {\tiny{5}}\\ \hline \hline
 {\color{olive}\tiny{\#$\uparrow$}} & {\tiny{1}}& {\tiny{1}}& {\tiny{3}}& {\tiny{0}}& {\tiny{7}}& {\tiny{3}}& {\tiny{6}}& {\tiny{4}}& {\tiny{0}}& {\tiny{0}}& {\tiny{3}}\\ \hline
 {\color{red}\tiny{\#$\downarrow$}} & {\tiny{7}}& {\tiny{6}}& {\tiny{3}}& {\tiny{3}}& {\tiny{0}}& {\tiny{4}}& {\tiny{1}}& {\tiny{2}}& {\tiny{3}}& {\tiny{5}}& {\tiny{0}}\\ \hline \hline
Average & 95.0& 94.77& 93.4& 87.72& 56.67& 72.48& 74.25& 67.19& 71.84& 58.65& 47.45& & \\ \hline 
All & {\color{red}\scriptsize{94.95} $\downarrow$ }& {\color{red}\scriptsize{94.42} $\downarrow$ }& \tiny{93.64}& {\color{red}\scriptsize{86.8} $\downarrow$ }& {\color{olive}\scriptsize{61.97} $\uparrow$ }& {\color{red}\scriptsize{71.72} $\downarrow$ }& {\color{olive}\scriptsize{74.36} $\uparrow$ }& {\color{olive}\scriptsize{67.98} $\uparrow$ }& {\color{olive}\scriptsize{74.61} $\uparrow$ }& {\color{red}\scriptsize{58.14} $\downarrow$ }& {\color{olive}\scriptsize{51.31} $\uparrow$ }& {\tiny{5}} & {\tiny{5}}\\ \hline 
Oracle& {\color{olive}\scriptsize{95.57} $\uparrow$ }& {\color{olive}\scriptsize{95.21} $\uparrow$ }& {\color{olive}\scriptsize{94.07} $\uparrow$ }&  {\color{blue}\tiny{88.24}}& {\color{olive}\scriptsize{61.74} $\uparrow$ }& {\color{olive}\scriptsize{73.1} $\uparrow$ }& {\color{olive}\scriptsize{75.24} $\uparrow$ }& {\color{olive}\scriptsize{68.22} $\uparrow$ }&  {\color{blue}\tiny{72.71}}&  {\color{blue}\tiny{62.04}}& {\color{olive}\scriptsize{50.15} $\uparrow$ }& {\tiny{8}} & {\tiny{0}}\\ \hline
\end{tabular}
\vspace{-7pt}
\caption{\small F1 scores for \tedec. We compare STL setting (blue), with pairwise MTL (+$\langle$task$\rangle$), All, and Oracle. We test on each task in the columns. Beneficial settings are in green. Harmful setting are in red. The last two columns indicate how many tasks are helped or harmed by the task at that row.}\label{tPwTEDec}}
\vspace{-5pt}
\end{table*}
\begin{table*}[ht]
\centering
\scriptsize{
\begin{tabular}{c|c|c|c|c|c|c|c|c|c|c|c|c|c}
 & \tinytask{upos} & \tinytask{xpos} & \tinytask{chunk} & \tinytask{ner} & \tinytask{mwe} & \tinytask{sem} & \tinytask{semtr} & \tinytask{supsense} & \tinytask{com} & \tinytask{frame} & \tinytask{hyp} & {\color{olive}\tiny{\#$\uparrow$}}  & {\color{red}\tiny{\#$\downarrow$}} \\ \hline
+\tinytask{upos} & {\color{blue}\scriptsize{95.4}}& \tiny{94.94}& {\color{olive}\scriptsize{94.0} $\uparrow$ }& {\color{red}\scriptsize{87.43} $\downarrow$ }& {\color{olive}\scriptsize{57.61} $\uparrow$ }& {\color{olive}\scriptsize{73.11} $\uparrow$ }& {\color{olive}\scriptsize{74.85} $\uparrow$ }& {\color{olive}\scriptsize{67.76} $\uparrow$ }& \tiny{72.09}& \tiny{62.27}& \tiny{48.27}& {\tiny{5}} & {\tiny{1}}\\ \hline
+\tinytask{xpos} & \tiny{95.42}& {\color{blue}\scriptsize{95.04}}& {\color{olive}\scriptsize{93.98} $\uparrow$ }& {\color{red}\scriptsize{87.71} $\downarrow$ }& {\color{olive}\scriptsize{58.26} $\uparrow$ }& \tiny{73.04}& {\color{olive}\scriptsize{74.66} $\uparrow$ }& {\color{olive}\scriptsize{67.77} $\uparrow$ }& \tiny{72.41}& \tiny{61.62}& {\color{olive}\scriptsize{48.06} $\uparrow$ }& {\tiny{5}} & {\tiny{1}}\\ \hline
+\tinytask{chunk} & \tiny{95.4}& \tiny{95.1}& {\color{blue}\scriptsize{93.49}}& \tiny{88.07}& {\color{olive}\scriptsize{58.06} $\uparrow$ }& {\color{olive}\scriptsize{73.13} $\uparrow$ }& {\color{olive}\scriptsize{74.77} $\uparrow$ }& \tiny{67.36}& \tiny{72.88}& \tiny{62.98}& \tiny{47.13}& {\tiny{3}} & {\tiny{0}}\\ \hline
+\tinytask{ner} & \tiny{95.29}& \tiny{95.05}& \tiny{93.54}& {\color{blue}\scriptsize{88.24}}& \tiny{53.4}& \tiny{72.91}& \tiny{74.04}& {\color{olive}\scriptsize{67.57} $\uparrow$ }& {\color{red}\scriptsize{70.78} $\downarrow$ }& \tiny{63.02}& \tiny{48.64}& {\tiny{1}} & {\tiny{1}}\\ \hline
+\tinytask{mwe} & {\color{red}\scriptsize{95.05} $\downarrow$ }& {\color{red}\scriptsize{94.66} $\downarrow$ }& \tiny{93.33}& \tiny{88.02}& {\color{blue}\scriptsize{53.07}}& \tiny{72.83}& {\color{olive}\scriptsize{74.66} $\uparrow$ }& \tiny{66.26}& \tiny{71.36}& \tiny{60.61}& \tiny{46.71}& {\tiny{1}} & {\tiny{2}}\\ \hline
+\tinytask{sem} & \tiny{95.27}& \tiny{94.93}& \tiny{93.52}& {\color{red}\scriptsize{87.49} $\downarrow$ }& {\color{olive}\scriptsize{58.62} $\uparrow$ }& {\color{blue}\scriptsize{72.77}}& {\color{olive}\scriptsize{74.41} $\uparrow$ }& {\color{olive}\scriptsize{68.1} $\uparrow$ }& \tiny{72.25}& \tiny{62.17}& \tiny{47.12}& {\tiny{3}} & {\tiny{1}}\\ \hline
+\tinytask{semtr} & \tiny{95.23}& \tiny{94.97}& \tiny{93.45}& {\color{red}\scriptsize{87.29} $\downarrow$ }& {\color{olive}\scriptsize{58.31} $\uparrow$ }& {\color{red}\scriptsize{72.17} $\downarrow$ }& {\color{blue}\scriptsize{74.02}}& \tiny{67.64}& \tiny{72.15}& \tiny{62.79}& \tiny{46.1}& {\tiny{1}} & {\tiny{2}}\\ \hline
+\tinytask{supsense} & \tiny{95.27}& \tiny{95.0}& {\color{red}\scriptsize{93.13} $\downarrow$ }& \tiny{87.92}& {\color{olive}\scriptsize{58.05} $\uparrow$ }& {\color{olive}\scriptsize{73.09} $\uparrow$ }& {\color{olive}\scriptsize{74.94} $\uparrow$ }& {\color{blue}\scriptsize{66.81}}& \tiny{72.12}& \tiny{61.96}& \tiny{47.24}& {\tiny{3}} & {\tiny{1}}\\ \hline
+\tinytask{com} & {\color{red}\scriptsize{93.6} $\downarrow$ }& {\color{red}\scriptsize{93.12} $\downarrow$ }& {\color{red}\scriptsize{91.86} $\downarrow$ }& {\color{red}\scriptsize{86.75} $\downarrow$ }& \tiny{51.71}& {\color{red}\scriptsize{70.18} $\downarrow$ }& {\color{red}\scriptsize{71.35} $\downarrow$ }& {\color{red}\scriptsize{65.55} $\downarrow$ }& {\color{blue}\scriptsize{72.71}}& \tiny{57.65}& \tiny{47.81}& {\tiny{0}} & {\tiny{7}}\\ \hline
+\tinytask{frame} & {\color{red}\scriptsize{95.0} $\downarrow$ }& {\color{red}\scriptsize{94.55} $\downarrow$ }& \tiny{93.29}& \tiny{87.99}& \tiny{53.3}& \tiny{72.49}& {\color{olive}\scriptsize{74.63} $\uparrow$ }& \tiny{66.75}& \tiny{72.1}& {\color{blue}\scriptsize{62.04}}& \tiny{46.66}& {\tiny{1}} & {\tiny{2}}\\ \hline
+\tinytask{hyp} & {\color{red}\scriptsize{94.43} $\downarrow$ }& {\color{red}\scriptsize{94.26} $\downarrow$ }& {\color{red}\scriptsize{93.13} $\downarrow$ }& \tiny{87.82}& \tiny{52.59}& \tiny{71.95}& \tiny{74.14}& \tiny{66.16}& \tiny{72.79}& \tiny{61.14}& {\color{blue}\scriptsize{46.73}}& {\tiny{0}} & {\tiny{3}}\\ \hline \hline
 {\color{olive}\tiny{\#$\uparrow$}} & {\tiny{0}}& {\tiny{0}}& {\tiny{2}}& {\tiny{0}}& {\tiny{6}}& {\tiny{3}}& {\tiny{7}}& {\tiny{4}}& {\tiny{0}}& {\tiny{0}}& {\tiny{1}}\\ \hline
 {\color{red}\tiny{\#$\downarrow$}} & {\tiny{4}}& {\tiny{4}}& {\tiny{3}}& {\tiny{5}}& {\tiny{0}}& {\tiny{2}}& {\tiny{1}}& {\tiny{1}}& {\tiny{1}}& {\tiny{0}}& {\tiny{0}}\\ \hline \hline
Average & 95.0& 94.66& 93.32& 87.65& 55.99& 72.49& 74.24& 67.09& 72.09& 61.62& 47.37& & \\ \hline 
All & {\color{red}\scriptsize{94.94} $\downarrow$ }& {\color{red}\scriptsize{94.3} $\downarrow$ }& {\color{olive}\scriptsize{93.7} $\uparrow$ }& {\color{red}\scriptsize{86.01} $\downarrow$ }& {\color{olive}\scriptsize{59.57} $\uparrow$ }& {\color{red}\scriptsize{71.58} $\downarrow$ }& \tiny{74.35}& {\color{olive}\scriptsize{68.02} $\uparrow$ }& {\color{olive}\scriptsize{74.61} $\uparrow$ }& \tiny{61.83}& {\color{olive}\scriptsize{49.5} $\uparrow$ }& {\tiny{5}} & {\tiny{4}}\\ \hline 
Oracle&  {\color{blue}\tiny{95.4}}&  {\color{blue}\tiny{95.04}}& {\color{olive}\scriptsize{93.93} $\uparrow$ }&  {\color{blue}\tiny{88.24}}& {\color{olive}\scriptsize{61.92} $\uparrow$ }& {\color{olive}\scriptsize{73.14} $\uparrow$ }& {\color{olive}\scriptsize{75.09} $\uparrow$ }& {\color{olive}\scriptsize{69.04} $\uparrow$ }&  {\color{blue}\tiny{72.71}}&  {\color{blue}\tiny{62.04}}& {\color{olive}\scriptsize{48.06} $\uparrow$ }& {\tiny{6}} & {\tiny{0}}\\ \hline
\end{tabular}
\vspace{-7pt}
\caption{\small F1 scores for \teenc. We compare STL setting (blue), with pairwise MTL (+$\langle$task$\rangle$), All, and Oracle. We test on each task in the columns. Beneficial settings are in green. Harmful setting are in red. The last two columns indicate how many tasks are helped or harmed by the task at that row.}\label{tPwTEEnc}}
\vspace{-5pt}
\end{table*}


\subsection{All MTL Results}

In addition to pairwise MTL results, we report the performances in the All and Oracle MTL settings in the last two rows of Table~\ref{tPwMultiDec}-\ref{tPwTEEnc}. We find that their performances depend largely on the trend in their corresponding pairwise MTL. We provide examples and discussion of such observations below.  

\begin{table*}[ht]
\centering
\scriptsize{
\begin{tabular}{c|c|c|c|c|c|c|c|c|c|c|c|c|c}
 & \tinytask{upos} & \tinytask{xpos} & \tinytask{chunk} & \tinytask{ner} & \tinytask{mwe} & \tinytask{sem} & \tinytask{semtr} & \tinytask{supsense} & \tinytask{com} & \tinytask{frame} & \tinytask{hyp} & {\color{olive}\tiny{\#$\uparrow$}}  & {\color{red}\tiny{\#$\downarrow$}} \\ \hline
All & {\scriptsize{95.04}  }& {\scriptsize{94.31}  }& \scriptsize{93.44}& {\scriptsize{86.38}  }& {\scriptsize{61.43}  }& {\scriptsize{71.53}  }& {\scriptsize{74.26}  }& {\scriptsize{68.1}  }& \scriptsize{74.54}& \scriptsize{59.71}& {\scriptsize{51.41}  }& & \\ \hline \hline 
\tiny{All -} \tinytask{upos} & & \tiny{94.03}& \tiny{93.59}& \tiny{86.03}& \tiny{61.28}& \tiny{70.87}& \tiny{73.54}& \tiny{68.27}& \tiny{74.42}& \tiny{58.47}& \tiny{51.13}& {\tiny{0}} & {\tiny{0}}\\ \hline
\tiny{All -} \tinytask{xpos} & {\color{red}\scriptsize{94.57} $\downarrow$ }& & \tiny{93.57}& \tiny{86.04}& \tiny{61.91}& \tiny{71.12}& \tiny{74.03}& \tiny{67.99}& \tiny{74.36}& \tiny{60.16}& \tiny{51.65}& {\tiny{0}} & {\tiny{1}}\\ \hline
\tiny{All -} \tinytask{chunk} & {\color{red}\scriptsize{94.84} $\downarrow$ }& \tiny{94.46}& & \tiny{86.05}& \tiny{61.01}& \tiny{71.07}& \tiny{73.97}& \tiny{68.26}& \tiny{74.2}& \tiny{60.01}& \tiny{50.27}& {\tiny{0}} & {\tiny{1}}\\ \hline
\tiny{All -} \tinytask{ner} & {\color{red}\scriptsize{94.81} $\downarrow$ }& \tiny{94.3}& \tiny{93.59}& & \tiny{62.69}& \tiny{70.82}& {\color{red}\scriptsize{73.51} $\downarrow$ }& \tiny{68.16}& \tiny{74.08}& \tiny{59.17}& \tiny{50.86}& {\tiny{0}} & {\tiny{2}}\\ \hline
\tiny{All -} \tinytask{mwe} & {\color{red}\scriptsize{94.93} $\downarrow$ }& \tiny{94.45}& \tiny{93.71}& \tiny{86.21}& & \tiny{71.01}& {\color{red}\scriptsize{73.61} $\downarrow$ }& \tiny{68.18}& \tiny{74.7}& \tiny{59.23}& \tiny{50.83}& {\tiny{0}} & {\tiny{2}}\\ \hline
\tiny{All -} \tinytask{sem} & \tiny{94.82}& \tiny{94.34}& \tiny{93.63}& \tiny{85.81}& \tiny{61.17}& & {\color{red}\scriptsize{71.97} $\downarrow$ }& \tiny{67.36}& \tiny{74.31}& \tiny{58.73}& \tiny{50.93}& {\tiny{0}} & {\tiny{1}}\\ \hline
\tiny{All -} \tinytask{semtr} & \tiny{94.83}& \tiny{94.35}& \tiny{93.58}& \tiny{86.11}& \tiny{63.04}& {\color{red}\scriptsize{69.72} $\downarrow$ }& & \tiny{68.17}& \tiny{74.2}& \tiny{59.49}& \tiny{51.27}& {\tiny{0}} & {\tiny{1}}\\ \hline
\tiny{All -} \tinytask{supsense} & \tiny{94.97}& \tiny{94.54}& \tiny{93.67}& \tiny{86.43}& \tiny{60.51}& \tiny{71.22}& {\color{red}\scriptsize{73.86} $\downarrow$ }& & \tiny{74.24}& \tiny{59.23}& \tiny{50.86}& {\tiny{0}} & {\tiny{1}}\\ \hline
\tiny{All -} \tinytask{com} & {\color{olive}\scriptsize{95.19} $\uparrow$ }& {\color{olive}\scriptsize{94.69} $\uparrow$ }& \tiny{93.67}& \tiny{86.6}& \tiny{61.95}& {\color{olive}\scriptsize{72.38} $\uparrow$ }& {\color{olive}\scriptsize{74.75} $\uparrow$ }& \tiny{68.67}& & {\color{olive}\scriptsize{62.37} $\uparrow$ }& \tiny{50.28}& {\tiny{5}} & {\tiny{0}}\\ \hline
\tiny{All -} \tinytask{frame} & \tiny{95.15}& \tiny{94.57}& \tiny{93.7}& \tiny{85.9}& \tiny{62.62}& \tiny{71.48}& \tiny{74.24}& \tiny{68.47}& \tiny{75.03}& & \tiny{50.89}& {\tiny{0}} & {\tiny{0}}\\ \hline
\tiny{All -} \tinytask{hyp} & \tiny{94.93}& \tiny{94.53}& {\color{olive}\scriptsize{93.78} $\uparrow$ }& \tiny{86.31}& \tiny{62.04}& \tiny{71.22}& \tiny{74.02}& \tiny{68.46}& \tiny{74.62}& \tiny{59.69}& & {\tiny{1}} & {\tiny{0}}\\ \hline \hline
 {\color{olive}\tiny{\#$\uparrow$}} & {\tiny{1}}& {\tiny{1}}& {\tiny{1}}& {\tiny{0}}& {\tiny{0}}& {\tiny{1}}& {\tiny{1}}& {\tiny{0}}& {\tiny{0}}& {\tiny{1}}& {\tiny{0}}\\ \hline
 {\color{red}\tiny{\#$\downarrow$}} & {\tiny{4}}& {\tiny{0}}& {\tiny{0}}& {\tiny{0}}& {\tiny{0}}& {\tiny{1}}& {\tiny{4}}& {\tiny{0}}& {\tiny{0}}& {\tiny{0}}& {\tiny{0}}\\ \hline 
\end{tabular}
\vspace{-7pt}
\caption{\small F1 scores for \multidec. We compare All with All-but-one settings (All - \btask{$\langle$task$\rangle$}).  We test on each task in the columns.  Beneficial settings are in green. Harmful setting are in red. }\label{tAllminusMultiDec}}
\vspace{-5pt}
\end{table*}

\paragraph{How much is STL vs. Pairwise MTL predictive of STL vs. All MTL?}
We find that the performance of pairwise MTL is predictive of the performance of All MTL to some degree.
Below we discuss the results in more detail. Note that we would like to be predictive in both the performance direction and magnitude (whether and how much the scores will improve or degrade over the baseline).

\emph{When pairwise MTL improves upon STL even slightly, All improves upon STL in all cases} (\btask{mwe}, \btask{semtr}, \btask{supsense}, and \btask{hyp}). This is despite the fact that jointly learning some pairs of tasks lead to performance degradation (\btask{com} and \btask{frame} in the case of \btask{supsense} and \btask{com} in the case of \btask{semtr}). Furthermore, when pairwise MTL leads to improvement in all cases (all pairwise rows in \btask{mwe} and \btask{hyp}), All MTL will achieve even better performance, suggesting that tasks are beneficial in a complementary manner and there is an advantage of MTL beyond two tasks. 

\emph{The opposite is almost true.} When pairwise MTL does not improve upon STL, most of the time All MTL will not improve upon STL, either --- with one exception: \btask{com}. Specifically, the pairwise MTL performances of \btask{upos}, \btask{xpos}, \btask{ner} and \btask{frame} (\tedec) are mostly negative and so are their All MTL performances. Furthermore, tasks can also be harmful in a complementary manner. For instance, in the case of \btask{ner}, All MTL achieves the lowest or the second lowest score when compared to any row of the pairwise MTL settings. In addition, \btask{sem}'s pairwise MTL performances are mixed, making the average score about the same or slightly worse than STL. However, the performance of All MTL when tested on \btask{sem} almost achieves the lowest. In other words, \btask{sem} is harmed more than it is benefited but pairwise MTL performances cannot tell. This suggests that harmful tasks are complementary while beneficial tasks are not.

Our results when \emph{tested on \btask{com} are the most surprising}. While none of pairwise MTL settings help (with some hurting), the performance of All MTL goes in the opposite direction, outperforming that of STL. Further characterization of task interaction is needed to reveal why this happens. One hypothesis is that instances in \btask{com} that are benefited by one task may be harmed by another. The joint training of all tasks thus works because tasks \emph{regularize} each other.  

We believe that our results open the doors to other interesting research questions. While the pairwise MTL performance is somewhat predictive of the performance direction of All MTL (except \btask{com}), the magnitude of that direction is difficult to predict. It is clear that additional factors beyond pairwise performance contribute to the success or failure of the All MTL setting. It would be useful to automatically identify these factors or design a metric to capture that. There have been initial attempts along this research direction in \cite{AlonsoP17,BingelS17,Bjerva17}, in which manually-defined task characteristics are found to be predictive of \emph{pairwise} MTL's failure or success.

\paragraph{Oracle MTL}

Recall that a task has an ``Oracle" set when the task is benefited from some other tasks according to its pairwise results (cf. Sect.~\ref{sTrainSettings}).
In general, our simple heuristic works well. Out of 20 cases where Oracle MTL performances exist, 16 are better than the performance of All MTL. In \btask{sem}, \btask{upos} and \btask{xpos} (\tedec, Oracle MTL is able to reverse the negative results obtained by All MTL to the positive ones, leading to improved scores over STL in all cases. This suggests that pairwise MTL performances are valuable knowledge if we want to go beyond two tasks. But, as mentioned previously, pairwise performance information fails in the case of \btask{com}; All MTL leads to improvement but we do not have an Oracle set in this case. 

Out of 4 cases where Oracle MTL does not improve upon All MTL, 3 is when we test on \btask{hyp} and one is when we test on \btask{mwe}. These two tasks are not harmed by any tasks. This result seems to suggest that sometimes ``neutral" tasks can help in MTL (but not always, for example, in \multidec and \teenc of \btask{mwe}). This also raises the question of whether there is a more effective way to construct an oracle set.

\subsection{Analysis}
\label{sAnalysis}

\paragraph{Task Contribution in All MTL}
How much does one particular task contribute to the performance of All MTL? To investigate this, we remove one task at a time and train the rest jointly. Results are shown in Table~\ref{tAllminusMultiDec} for the method \multidec -- results for other two methods are in Appendix~\ref{sSuppAllButOne} as they are similar to \multidec qualitatively. 
We find that \btask{upos}, \btask{sem} and \btask{semtr} are in general sensitive to a task being removed from All MTL.
Moreover, at least one task significantly contributes to the success of All MTL at some point; if we remove it, the performance will drop.
On the other hand, \btask{com} generally negatively affects the performance of All MTL as removing it often leads to performance improvement.

\paragraph{Task Embeddings}

\begin{figure*}[ht]
\centering
\includegraphics[width=0.3\textwidth]{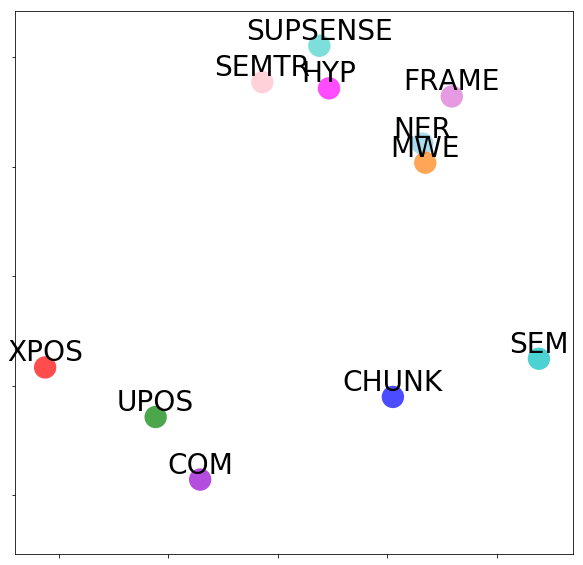}
\vspace{-6pt}
\caption{\small t-SNE visualization of the embeddings of the 11 tasks that are learned from \tedec} \label{fTaskEmb}
\vspace{-8pt}
\end{figure*}

Fig.~\ref{fTaskEmb} shows t-SNE visualization \cite{tSNE} of task embeddings learned from \tedec\footnote{We observed that task embeddings learned from \teenc are not consistent across multiple runs.} in the All MTL setting.
The learned task embeddings reflect our knowledge about similarities between tasks, where there are clusters of syntactic and semantic tasks. We also learn that sentence compression (\btask{com}) is more syntactic, whereas multi-word expression identification (\btask{mwe}) and hyper-text detection (\btask{hyp}) are more semantic. Interestingly, \btask{chunk} seems to be in between, which may explain why it never harms any tasks in any settings (cf. Sect.~\ref{sExpPwRes}).

In general, it is not obvious how to translate task similarities derived from task embeddings into something indicative of MTL performance. While our task embeddings could be considered as ``task characteristics" vectors, they are not guaranteed to be interpretable. We thus leave a thorough investigation of information captured by task embeddings to future work. 

Nevertheless, we observe that task embeddings disentangle ``sentences/tags" and ``actual task" to some degree. For instance, if we consider the locations of each pair of tasks that use the same set of sentences for training in Fig.~\ref{fTaskEmb}, we see that \btask{sem} and \btask{semtr} (or \btask{mwe} and \btask{supsense}) are not neighbors, while \btask{xpos} and \btask{upos} are. On the other hand, \btask{mwe} and \btask{ner} are neighbors, even though their label set size and entropy are not the closest. These observations suggest that hand-designed task features used in \cite{BingelS17} may not be the most informative characterization for predicting MTL performance.

\section{Related Work}
\label{sRelated}

For a comprehensive overview of MTL in NLP, see Chapter 20 of \cite{Goldberg17} and \cite{Ruder17}. Here we highlight those which are mostly relevant.

MTL for NLP has been popular since a unified architecture was proposed by \cite{CollobertW08,CollobertWBKKK11}. As for sequence to sequence learning \cite{SutskeverVL14}, general multi-task learning frameworks are explored by \cite{LuongLSVK16}.

Our work is different from existing work in several aspects.
First, the majority of the work focuses on two tasks, often with one being the main task and the other being the auxiliary one~\cite{SogaardG16,BjervaPB16,PlankSG16,AlonsoP17,BingelS17}.
For example, \btask{pos} is the auxiliary task in \cite{SogaardG16} while \btask{chunk}, CCG supertagging (\btask{ccg}) \cite{Clark02}, \btask{ner}, \btask{sem}, or \btask{mwe}+\btask{supsense} is the main one. They find that \btask{pos} benefits \btask{chunk} and \btask{ccg}. 
Another line of work considers language modeling as the auxiliary objective \cite{GodwinSR16,Rei17,LiuSXRGPH18}.
Besides sequence tagging, some work considers two high-level tasks or one high-level task with another lower-level one. Examples are dependency parsing (\btask{dep}) with \btask{pos} \cite{ZhangW16}, with \btask{mwe} \cite{ConstantN16}, or with semantic role labeling (\btask{srl}) \cite{ShiTZ16}; machine translation (\btask{translate}) with \btask{pos} or \btask{dep} \cite{NiehuesC17,EriguchiTC17}; sentence extraction and \btask{com} \cite{MartinsS09,BergKirkpatrickGK11,AlmeidaM13}.

Exceptions to this include the work of \cite{CollobertWBKKK11}, which considers four tasks: \btask{pos}, \btask{chunk}, \btask{ner}, and \btask{srl}; \cite{RaganatoBN17}, which considers three: word sense disambiguation with \btask{pos} and coarse-grained semantic tagging based on WordNet lexicographer files; \cite{HashimotoXTS17}, which considers five: \btask{pos}, \btask{chunk}, \btask{dep}, semantic relatedness, and textual entailment; \cite{NiehuesC17,KiperwasserB18}, which both consider three: \btask{translate} with \btask{pos} and \btask{ner}, and \btask{translate} with \btask{pos} and \btask{dep}, respectively. We consider as many as 11 tasks jointly.

Second, we choose to focus on model architectures that are generic enough to be shared by many tasks. Our structure is similar to \cite{CollobertWBKKK11}, but we also explore frameworks related to task embeddings and propose two variants. In contrast, recent work considers stacked architectures (mostly for sequence tagging) in which tasks can supervise at different layers of a network \cite{SogaardG16,KlerkeGS16,PlankSG16,AlonsoP17,BingelS17,HashimotoXTS17}. More complicated structures require more sophisticated MTL methods when the number of tasks grows and thus prevent us from concentrating on analyzing relationships among tasks. For this reason, we leave MTL for complicated models for future work.

The purpose of our study is relevant to but different from transfer learning, where the setting designates one or more target tasks and focuses on whether the target tasks can be learned more effectively from the source tasks; see e.g., \cite{MouMYLXZJ16,YangSC17}.  
\section{Discussion and Future Work} \label{sDis}

We conduct an empirical study on MTL for sequence tagging, which so far has been mostly studied with two or a few tasks. We also propose two alternative frameworks that augment taggers with task embeddings. Our results provide insights regarding task relatedness and show benefits of the MTL approaches. Nevertheless, we believe that our work simply scratches the surface of MTL. The characterization of task relationships seems to go beyond the performances of pairwise MTL training or similarities of their task embeddings. We are also interested in exploring further other techniques to MTL, especially when tasks become more complicated. For example, it is not clear how to best represent task specification as well as how to incorporate them into NLP systems. Finally, the definition of tasks can be relaxed to include domains or languages. Combining all these will move us toward the goal of having a \emph{single} robust, generalizable NLP agent that is equipped with a diverse set of skills.

\paragraph{Acknowledgments} {This work is partially supported by USC Graduate Fellowship, NSF IIS-1065243, 1451412, 1513966/1632803/1833137, 1208500, CCF-1139148, a Google Research Award, an Alfred P. Sloan Research Fellowship, gifts from Facebook and Netflix, and ARO\# W911NF-12-1-0241 and W911NF-15-1-0484.

\bibliographystyle{acl}
\bibliography{main_supp}

\newpage
\appendix

\section{Comparison between different MTL approaches}
\label{sSuppCompareApp}

\begin{table*}[ht]
\centering
\scriptsize{
\begin{tabular}{c|c|c|c|c|c|c|c|c|c|c|c|c||c}
 Settings & Method & \tinytask{upos} & \tinytask{xpos} & \tinytask{chunk} & \tinytask{ner} & \tinytask{mwe} & \tinytask{sem} & \tinytask{semtr} & \tinytask{supsense} & \tinytask{com} & \tinytask{frame} & \tinytask{hyp} & {\scriptsize{Average}}\\ \hline
\multicolumn{2}{c|}{STL}& {\tiny{95.4}}& {\tiny{95.04}}& {\tiny{93.49}}& {\tiny{88.24}}& {\tiny{53.07}}& {\tiny{72.77}}& {\tiny{74.02}}& {\tiny{66.81}}& {\tiny{72.71}}& {\tiny{62.04}}& {\tiny{46.73}}& {\tiny{74.58}}\\ \hline \hline
(Average)& \multidec& {\tiny{94.97}}& {\tiny{94.65}}& {\tiny{93.37}}& {\tiny{87.67}}& {\tiny{57.21}}& {\tiny{72.63}}& {\tiny{74.38}}& {\tiny{67.39}}& {\tiny{72.12}}& {\tiny{61.3}}& {\tiny{47.99}}& {\tiny{74.88}}\\ \cline{2-14}
Pairwise& \tedec& {\tiny{95.0}}& {\tiny{94.77}}& {\tiny{93.4}}& {\tiny{87.72}}& {\tiny{56.67}}& {\tiny{72.48}}& {\tiny{74.25}}& {\tiny{67.19}}& {\tiny{71.84}}& {\tiny{58.65}}& {\tiny{47.45}}& {\tiny{74.49}}\\ \cline{2-14}
& \teenc& {\tiny{95.0}}& {\tiny{94.66}}& {\tiny{93.32}}& {\tiny{87.65}}& {\tiny{55.99}}& {\tiny{72.49}}& {\tiny{74.24}}& {\tiny{67.09}}& {\tiny{72.09}}& {\tiny{61.62}}& {\tiny{47.37}}& {\tiny{74.68}}\\ \hline \hline
& \multidec& {\tiny{95.04}}& {\tiny{94.31}}& {\tiny{93.44}}& {\tiny{86.38}}& {\tiny{61.43}}& {\tiny{71.53}}& {\tiny{74.26}}& {\tiny{68.1}}& {\tiny{74.54}}& {\tiny{59.71}}& {\tiny{51.41}}& {\tiny{75.47}}\\ \cline{2-14}
All& \tedec& {\tiny{94.95}}& {\tiny{94.42}}& {\tiny{93.64}}& {\tiny{86.8}}& {\tiny{61.97}}& {\tiny{71.72}}& {\tiny{74.36}}& {\tiny{67.98}}& {\tiny{74.61}}& {\tiny{58.14}}& {\tiny{51.31}}& {\tiny{75.44}}\\ \cline{2-14}
& \teenc& {\tiny{94.94}}& {\tiny{94.3}}& {\tiny{93.7}}& {\tiny{86.01}}& {\tiny{59.57}}& {\tiny{71.58}}& {\tiny{74.35}}& {\tiny{68.02}}& {\tiny{74.61}}& {\tiny{61.83}}& {\tiny{49.5}}& {\tiny{75.31}}\\ \hline \hline
(Average)& \multidec& {\tiny{94.91}}& {\tiny{94.43}}& {\tiny{93.65}}& {\tiny{86.15}}& {\tiny{61.82}}& {\tiny{71.09}}& {\tiny{73.75}}& {\tiny{68.2}}& {\tiny{74.42}}& {\tiny{59.66}}& {\tiny{50.9}}& {\tiny{75.36}}\\ \cline{2-14}
All-but-one& \tedec& {\tiny{94.83}}& {\tiny{94.4}}& {\tiny{93.64}}& {\tiny{86.39}}& {\tiny{60.55}}& {\tiny{70.95}}& {\tiny{73.74}}& {\tiny{67.81}}& {\tiny{74.47}}& {\tiny{58.66}}& {\tiny{50.86}}& {\tiny{75.12}}\\ \cline{2-14}
& \teenc& {\tiny{94.77}}& {\tiny{94.35}}& {\tiny{93.53}}& {\tiny{85.96}}& {\tiny{60.23}}& {\tiny{70.83}}& {\tiny{73.64}}& {\tiny{68.15}}& {\tiny{74.05}}& {\tiny{61.23}}& {\tiny{50.15}}& {\tiny{75.17}}\\ \hline \hline
\end{tabular}
\vspace{-7pt}
\caption{\small Comparison between MTL approaches}\label{tCompareMtl}}
\vspace{-5pt}
\end{table*}

In Table~\ref{tCompareMtl}, we summarize the results of different MTL approaches. We observe no significant differences between those methods.

\section{Additional results on All-but-one settings}
\label{sSuppAllButOne}

Table~\ref{tAllminusTEDec} and Table~\ref{tAllminusTEEnc} compare All and All-but-one settings for \tedec and \teenc, respectively. We show similar results for \multidec in the main text. 

\begin{table*}[ht]
\centering
\scriptsize{
\begin{tabular}{c|c|c|c|c|c|c|c|c|c|c|c|c|c}
 & \tinytask{upos} & \tinytask{xpos} & \tinytask{chunk} & \tinytask{ner} & \tinytask{mwe} & \tinytask{sem} & \tinytask{semtr} & \tinytask{supsense} & \tinytask{com} & \tinytask{frame} & \tinytask{hyp} & {\color{olive}\tiny{\#$\uparrow$}}  & {\color{red}\tiny{\#$\downarrow$}} \\ \hline
All & {\scriptsize{94.95}  }& {\scriptsize{94.42}  }& \scriptsize{93.64}& {\scriptsize{86.8}  }& {\scriptsize{61.97}  }& {\scriptsize{71.72}  }& {\scriptsize{74.36}  }& {\scriptsize{67.98}  }& {\scriptsize{74.61}  }& {\scriptsize{58.14}  }& {\scriptsize{51.31}  }& & \\ \hline \hline 
\tiny{All -} \tinytask{upos} & & {\color{red}\scriptsize{94.06} $\downarrow$ }& \tiny{93.44}& \tiny{86.47}& \tiny{60.48}& {\color{red}\scriptsize{71.08} $\downarrow$ }& \tiny{73.79}& \tiny{68.1}& \tiny{74.69}& \tiny{58.32}& \tiny{50.83}& {\tiny{0}} & {\tiny{2}}\\ \hline
\tiny{All -} \tinytask{xpos} & {\color{red}\scriptsize{94.38} $\downarrow$ }& & \tiny{93.6}& \tiny{86.68}& \tiny{60.09}& {\color{red}\scriptsize{70.98} $\downarrow$ }& {\color{red}\scriptsize{73.78} $\downarrow$ }& \tiny{67.9}& \tiny{74.26}& \tiny{58.31}& \tiny{50.6}& {\tiny{0}} & {\tiny{3}}\\ \hline
\tiny{All -} \tinytask{chunk} & {\color{red}\scriptsize{94.6} $\downarrow$ }& \tiny{94.29}& & \tiny{86.08}& \tiny{60.6}& {\color{red}\scriptsize{70.39} $\downarrow$ }& {\color{red}\scriptsize{73.36} $\downarrow$ }& \tiny{68.07}& \tiny{74.47}& \tiny{58.73}& \tiny{51.1}& {\tiny{0}} & {\tiny{3}}\\ \hline
\tiny{All -} \tinytask{ner} & {\color{red}\scriptsize{94.69} $\downarrow$ }& \tiny{94.31}& \tiny{93.69}& & {\color{red}\scriptsize{60.48} $\downarrow$ }& {\color{red}\scriptsize{70.64} $\downarrow$ }& {\color{red}\scriptsize{73.59} $\downarrow$ }& \tiny{67.51}& \tiny{74.49}& \tiny{58.19}& \tiny{50.44}& {\tiny{0}} & {\tiny{4}}\\ \hline
\tiny{All -} \tinytask{mwe} & \tiny{94.93}& \tiny{94.46}& \tiny{93.72}& {\color{red}\scriptsize{86.21} $\downarrow$ }& & {\color{red}\scriptsize{71.11} $\downarrow$ }& \tiny{74.04}& \tiny{67.38}& \tiny{74.49}& \tiny{57.6}& \tiny{50.5}& {\tiny{0}} & {\tiny{2}}\\ \hline
\tiny{All -} \tinytask{sem} & \tiny{94.86}& \tiny{94.41}& \tiny{93.6}& {\color{red}\scriptsize{85.97} $\downarrow$ }& {\color{red}\scriptsize{59.94} $\downarrow$ }& & {\color{red}\scriptsize{72.26} $\downarrow$ }& \tiny{67.35}& \tiny{74.34}& \tiny{59.08}& \tiny{50.48}& {\tiny{0}} & {\tiny{3}}\\ \hline
\tiny{All -} \tinytask{semtr} & \tiny{94.8}& \tiny{94.28}& \tiny{93.56}& {\color{red}\scriptsize{86.23} $\downarrow$ }& \tiny{61.23}& {\color{red}\scriptsize{69.62} $\downarrow$ }& & \tiny{68.16}& \tiny{74.36}& \tiny{58.85}& \tiny{51.5}& {\tiny{0}} & {\tiny{2}}\\ \hline
\tiny{All -} \tinytask{supsense} & \tiny{94.82}& \tiny{94.4}& \tiny{93.67}& \tiny{86.49}& \tiny{59.11}& {\color{red}\scriptsize{71.02} $\downarrow$ }& {\color{red}\scriptsize{73.76} $\downarrow$ }& & \tiny{74.69}& \tiny{58.28}& \tiny{51.96}& {\tiny{0}} & {\tiny{2}}\\ \hline
\tiny{All -} \tinytask{com} & {\color{olive}\scriptsize{95.19} $\uparrow$ }& {\color{olive}\scriptsize{94.76} $\uparrow$ }& \tiny{93.79}& {\color{red}\scriptsize{86.25} $\downarrow$ }& \tiny{62.02}& \tiny{72.32}& {\color{olive}\scriptsize{74.92} $\uparrow$ }& \tiny{67.62}& & {\color{olive}\scriptsize{60.72} $\uparrow$ }& {\color{red}\scriptsize{50.0} $\downarrow$ }& {\tiny{4}} & {\tiny{2}}\\ \hline
\tiny{All -} \tinytask{frame} & \tiny{95.03}& \tiny{94.6}& \tiny{93.64}& \tiny{86.68}& {\color{red}\scriptsize{60.52} $\downarrow$ }& {\color{red}\scriptsize{71.11} $\downarrow$ }& \tiny{73.9}& \tiny{67.69}& \tiny{74.49}& & \tiny{51.23}& {\tiny{0}} & {\tiny{2}}\\ \hline
\tiny{All -} \tinytask{hyp} & \tiny{94.94}& \tiny{94.45}& \tiny{93.69}& \tiny{86.86}& \tiny{61.07}& \tiny{71.22}& {\color{red}\scriptsize{74.04} $\downarrow$ }& \tiny{68.32}& \tiny{74.4}& \tiny{58.55}& & {\tiny{0}} & {\tiny{1}}\\ \hline \hline
 {\color{olive}\tiny{\#$\uparrow$}} & {\tiny{1}}& {\tiny{1}}& {\tiny{0}}& {\tiny{0}}& {\tiny{0}}& {\tiny{0}}& {\tiny{1}}& {\tiny{0}}& {\tiny{0}}& {\tiny{1}}& {\tiny{0}}\\ \hline
 {\color{red}\tiny{\#$\downarrow$}} & {\tiny{3}}& {\tiny{1}}& {\tiny{0}}& {\tiny{4}}& {\tiny{3}}& {\tiny{8}}& {\tiny{6}}& {\tiny{0}}& {\tiny{0}}& {\tiny{0}}& {\tiny{1}}\\ \hline 
\end{tabular}
\vspace{-7pt}
\caption{\small F1 scores for \tedec. We compare All with All-but-one settings (All - \btask{$\langle$task$\rangle$}).  We test on each task in the columns.  Beneficial settings are in green. Harmful setting are in red. }\label{tAllminusTEDec}}
\vspace{-5pt}
\end{table*}
\begin{table*}[ht]
\centering
\scriptsize{
\begin{tabular}{c|c|c|c|c|c|c|c|c|c|c|c|c|c}
 & \tinytask{upos} & \tinytask{xpos} & \tinytask{chunk} & \tinytask{ner} & \tinytask{mwe} & \tinytask{sem} & \tinytask{semtr} & \tinytask{supsense} & \tinytask{com} & \tinytask{frame} & \tinytask{hyp} & {\color{olive}\tiny{\#$\uparrow$}}  & {\color{red}\tiny{\#$\downarrow$}} \\ \hline
All & {\scriptsize{94.94}  }& {\scriptsize{94.3}  }& {\scriptsize{93.7}  }& {\scriptsize{86.01}  }& {\scriptsize{59.57}  }& {\scriptsize{71.58}  }& \scriptsize{74.35}& {\scriptsize{68.02}  }& {\scriptsize{74.61}  }& \scriptsize{61.83}& {\scriptsize{49.5}  }& & \\ \hline \hline 
\tiny{All -} \tinytask{upos} & & \tiny{94.0}& {\color{red}\scriptsize{93.36} $\downarrow$ }& \tiny{85.98}& \tiny{59.58}& \tiny{70.68}& \tiny{73.66}& \tiny{68.19}& \tiny{74.07}& \tiny{60.51}& \tiny{50.23}& {\tiny{0}} & {\tiny{1}}\\ \hline
\tiny{All -} \tinytask{xpos} & {\color{red}\scriptsize{94.24} $\downarrow$ }& & {\color{red}\scriptsize{93.29} $\downarrow$ }& \tiny{85.8}& \tiny{59.81}& {\color{red}\scriptsize{70.57} $\downarrow$ }& {\color{red}\scriptsize{73.64} $\downarrow$ }& \tiny{68.47}& \tiny{73.94}& \tiny{60.13}& \tiny{50.39}& {\tiny{0}} & {\tiny{4}}\\ \hline
\tiny{All -} \tinytask{chunk} & \tiny{94.66}& \tiny{94.3}& & \tiny{85.73}& \tiny{61.58}& \tiny{70.78}& \tiny{73.65}& \tiny{67.87}& {\color{red}\scriptsize{73.67} $\downarrow$ }& \tiny{61.73}& \tiny{50.18}& {\tiny{0}} & {\tiny{1}}\\ \hline
\tiny{All -} \tinytask{ner} & \tiny{94.71}& \tiny{94.25}& \tiny{93.5}& & \tiny{59.05}& {\color{red}\scriptsize{70.58} $\downarrow$ }& {\color{red}\scriptsize{73.4} $\downarrow$ }& \tiny{67.95}& \tiny{74.16}& \tiny{59.96}& \tiny{49.95}& {\tiny{0}} & {\tiny{2}}\\ \hline
\tiny{All -} \tinytask{mwe} & \tiny{94.94}& \tiny{94.5}& \tiny{93.63}& \tiny{86.1}& & \tiny{71.12}& \tiny{73.75}& \tiny{69.0}& \tiny{74.28}& \tiny{61.51}& \tiny{49.81}& {\tiny{0}} & {\tiny{0}}\\ \hline
\tiny{All -} \tinytask{sem} & \tiny{94.76}& \tiny{94.32}& \tiny{93.45}& \tiny{85.58}& \tiny{59.47}& & {\color{red}\scriptsize{72.21} $\downarrow$ }& \tiny{67.77}& \tiny{74.2}& \tiny{61.76}& {\color{olive}\scriptsize{50.15} $\uparrow$ }& {\tiny{1}} & {\tiny{1}}\\ \hline
\tiny{All -} \tinytask{semtr} & \tiny{94.68}& \tiny{94.25}& \tiny{93.54}& \tiny{86.02}& \tiny{60.59}& {\color{red}\scriptsize{69.86} $\downarrow$ }& & \tiny{67.96}& {\color{red}\scriptsize{73.81} $\downarrow$ }& \tiny{61.31}& {\color{olive}\scriptsize{51.72} $\uparrow$ }& {\tiny{1}} & {\tiny{2}}\\ \hline
\tiny{All -} \tinytask{supsense} & \tiny{94.8}& \tiny{94.27}& \tiny{93.56}& \tiny{86.04}& \tiny{59.25}& {\color{red}\scriptsize{70.53} $\downarrow$ }& {\color{red}\scriptsize{73.27} $\downarrow$ }& & \tiny{74.3}& \tiny{59.98}& \tiny{50.01}& {\tiny{0}} & {\tiny{2}}\\ \hline
\tiny{All -} \tinytask{com} & {\color{olive}\scriptsize{95.25} $\uparrow$ }& {\color{olive}\scriptsize{94.72} $\uparrow$ }& \tiny{93.82}& \tiny{86.23}& \tiny{60.63}& {\color{olive}\scriptsize{72.38} $\uparrow$ }& {\color{olive}\scriptsize{75.06} $\uparrow$ }& \tiny{67.94}& & \tiny{63.55}& \tiny{48.77}& {\tiny{4}} & {\tiny{0}}\\ \hline
\tiny{All -} \tinytask{frame} & \tiny{94.84}& \tiny{94.39}& {\color{red}\scriptsize{93.51} $\downarrow$ }& \tiny{85.99}& \tiny{61.21}& \tiny{70.78}& \tiny{73.69}& \tiny{68.13}& \tiny{74.3}& & \tiny{50.35}& {\tiny{0}} & {\tiny{1}}\\ \hline
\tiny{All -} \tinytask{hyp} & \tiny{94.86}& \tiny{94.45}& \tiny{93.59}& \tiny{86.1}& \tiny{61.09}& {\color{red}\scriptsize{71.03} $\downarrow$ }& \tiny{74.09}& \tiny{68.17}& {\color{red}\scriptsize{73.78} $\downarrow$ }& \tiny{61.91}& & {\tiny{0}} & {\tiny{2}}\\ \hline \hline
 {\color{olive}\tiny{\#$\uparrow$}} & {\tiny{1}}& {\tiny{1}}& {\tiny{0}}& {\tiny{0}}& {\tiny{0}}& {\tiny{1}}& {\tiny{1}}& {\tiny{0}}& {\tiny{0}}& {\tiny{0}}& {\tiny{2}}\\ \hline
 {\color{red}\tiny{\#$\downarrow$}} & {\tiny{1}}& {\tiny{0}}& {\tiny{3}}& {\tiny{0}}& {\tiny{0}}& {\tiny{5}}& {\tiny{4}}& {\tiny{0}}& {\tiny{3}}& {\tiny{0}}& {\tiny{0}}\\ \hline 
\end{tabular}
\vspace{-7pt}
\caption{\small F1 scores for \teenc. We compare All with All-but-one settings (All - \btask{$\langle$task$\rangle$}).  We test on each task in the columns.  Beneficial settings are in green. Harmful setting are in red. }\label{tAllminusTEEnc}}
\vspace{-5pt}
\end{table*}

\section{Detailed results separated by the tasks being tested on}
\label{sSuppStd}
In Table~\ref{tMultiTaskuposuni}-\ref{tMultiTaskhyphyp}, we provide F1 scores with standard deviations in all settings.
Each table corresponds to a task we test our models on. Rows denote training settings and columns denote MTL approaches.
\begin{landscape}

\begin{table*}[t]
\centering
\parbox{.49\linewidth}{
\centering
\scriptsize{
\begin{tabular}{c|c|c|c|c}
\multicolumn{2}{c}{Trained with} & \multicolumn{3}{|c}{Tested on \task{upos}} \\ \cline{3-5}
 \multicolumn{2}{c|}{} & \multidec & \tedec & \teenc \\ \hline
\multicolumn{2}{c}{ \task{upos} only } & \multicolumn{3}{|c}{95.4\tiny{ $\pm$ 0.08}}\\ \hline
\parbox[t]{1mm}{\multirow{11}{*}{\rotatebox[origin = c]{90}{Pairwise}}}&\task{+xpos}&95.38\tiny{ $\pm$ 0.03}&95.4\tiny{ $\pm$ 0.04}&95.42\tiny{ $\pm$ 0.07} \\ 
&\task{+chunk}&95.43\tiny{ $\pm$ 0.11}&{\color{olive}95.57\tiny{ $\pm$ 0.02} $\uparrow$ }&95.4\tiny{ $\pm$ 0.0} \\ 
&\task{+ner}&95.38\tiny{ $\pm$ 0.1}&95.32\tiny{ $\pm$ 0.03}&95.29\tiny{ $\pm$ 0.04} \\ 
&\task{+mwe}&{\color{red}95.15\tiny{ $\pm$ 0.05} $\downarrow$ }&{\color{red}95.11\tiny{ $\pm$ 0.07} $\downarrow$ }&{\color{red}95.05\tiny{ $\pm$ 0.05} $\downarrow$ } \\ 
&\task{+sem}&95.23\tiny{ $\pm$ 0.14}&{\color{red}95.2\tiny{ $\pm$ 0.05} $\downarrow$ }&95.27\tiny{ $\pm$ 0.08} \\ 
&\task{+semtr}&95.17\tiny{ $\pm$ 0.15}&{\color{red}95.21\tiny{ $\pm$ 0.03} $\downarrow$ }&95.23\tiny{ $\pm$ 0.13} \\ 
&\task{+supsense}&{\color{red}95.08\tiny{ $\pm$ 0.08} $\downarrow$ }&{\color{red}95.05\tiny{ $\pm$ 0.04} $\downarrow$ }&95.27\tiny{ $\pm$ 0.08} \\ 
&\task{+com}&{\color{red}93.04\tiny{ $\pm$ 0.77} $\downarrow$ }&{\color{red}94.03\tiny{ $\pm$ 0.42} $\downarrow$ }&{\color{red}93.6\tiny{ $\pm$ 0.15} $\downarrow$ } \\ 
&\task{+frame}&{\color{red}94.98\tiny{ $\pm$ 0.13} $\downarrow$ }&{\color{red}94.79\tiny{ $\pm$ 0.09} $\downarrow$ }&{\color{red}95.0\tiny{ $\pm$ 0.07} $\downarrow$ } \\ 
&\task{+hyp}&{\color{red}94.84\tiny{ $\pm$ 0.07} $\downarrow$ }&{\color{red}94.35\tiny{ $\pm$ 0.21} $\downarrow$ }&{\color{red}94.43\tiny{ $\pm$ 0.15} $\downarrow$ } \\  \cline{2-5} 
& Average&94.97&95.0&95.0 \\ \hline 
\parbox[t]{1mm}{\multirow{10}{*}{\rotatebox[origin = c]{90}{All-but-one}}}&All - \task{xpos}&{\color{red}94.57\tiny{ $\pm$ 0.12} $\downarrow$ }&{\color{red}94.38\tiny{ $\pm$ 0.05} $\downarrow$ }&{\color{red}94.24\tiny{ $\pm$ 0.24} $\downarrow$ } \\ 
&All - \task{chunk}&{\color{red}94.84\tiny{ $\pm$ 0.01} $\downarrow$ }&{\color{red}94.6\tiny{ $\pm$ 0.1} $\downarrow$ }&{\color{red}94.66\tiny{ $\pm$ 0.15} $\downarrow$ } \\ 
&All - \task{ner}&{\color{red}94.81\tiny{ $\pm$ 0.07} $\downarrow$ }&{\color{red}94.69\tiny{ $\pm$ 0.05} $\downarrow$ }&{\color{red}94.71\tiny{ $\pm$ 0.07} $\downarrow$ } \\ 
&All - \task{mwe}&{\color{red}94.93\tiny{ $\pm$ 0.01} $\downarrow$ }&{\color{red}94.93\tiny{ $\pm$ 0.08} $\downarrow$ }&{\color{red}94.94\tiny{ $\pm$ 0.04} $\downarrow$ } \\ 
&All - \task{sem}&{\color{red}94.82\tiny{ $\pm$ 0.17} $\downarrow$ }&{\color{red}94.86\tiny{ $\pm$ 0.08} $\downarrow$ }&{\color{red}94.76\tiny{ $\pm$ 0.15} $\downarrow$ } \\ 
&All - \task{semtr}&{\color{red}94.83\tiny{ $\pm$ 0.12} $\downarrow$ }&{\color{red}94.8\tiny{ $\pm$ 0.03} $\downarrow$ }&{\color{red}94.68\tiny{ $\pm$ 0.17} $\downarrow$ } \\ 
&All - \task{supsense}&{\color{red}94.97\tiny{ $\pm$ 0.07} $\downarrow$ }&{\color{red}94.82\tiny{ $\pm$ 0.03} $\downarrow$ }&{\color{red}94.8\tiny{ $\pm$ 0.07} $\downarrow$ } \\ 
&All - \task{com}&{\color{red}95.19\tiny{ $\pm$ 0.05} $\downarrow$ }&{\color{red}95.19\tiny{ $\pm$ 0.04} $\downarrow$ }&{\color{red}95.25\tiny{ $\pm$ 0.02} $\downarrow$ } \\ 
&All - \task{frame}&{\color{red}95.15\tiny{ $\pm$ 0.07} $\downarrow$ }&95.03\tiny{ $\pm$ 0.17}&{\color{red}94.84\tiny{ $\pm$ 0.1} $\downarrow$ } \\ 
&All - \task{hyp}&{\color{red}94.93\tiny{ $\pm$ 0.18} $\downarrow$ }&{\color{red}94.94\tiny{ $\pm$ 0.11} $\downarrow$ }&{\color{red}94.86\tiny{ $\pm$ 0.04} $\downarrow$ } \\  \hline 
\multicolumn{2}{c|}{All}&{\color{red}95.04\tiny{ $\pm$ 0.03} $\downarrow$ }&{\color{red}94.95\tiny{ $\pm$ 0.08} $\downarrow$ }&{\color{red}94.94\tiny{ $\pm$ 0.1} $\downarrow$ } \\  \hline 
\multicolumn{2}{c|}{ Oracle } & {\color{blue}95.4 $\pm$ 0.08}&95.57\tiny{ $\pm$ 0.02}&{\color{blue}95.4 $\pm$ 0.08} \\ \hline
\end{tabular}
\caption{\small F1 score tested on the task \task{upos} in different training scenarios}\label{tMultiTaskuposuni}}}
\parbox{.49\linewidth}{
\centering
\scriptsize{
\begin{tabular}{c|c|c|c|c}
\multicolumn{2}{c}{Trained with} & \multicolumn{3}{|c}{Tested on \task{xpos}} \\ \cline{3-5}
 \multicolumn{2}{c|}{} & \multidec & \tedec & \teenc \\ \hline
\multicolumn{2}{c}{ \task{xpos} only } & \multicolumn{3}{|c}{95.04\tiny{ $\pm$ 0.06}}\\ \hline
\parbox[t]{1mm}{\multirow{11}{*}{\rotatebox[origin = c]{90}{Pairwise}}}&\task{+upos}&95.01\tiny{ $\pm$ 0.04}&94.99\tiny{ $\pm$ 0.03}&94.94\tiny{ $\pm$ 0.05} \\ 
&\task{+chunk}&95.1\tiny{ $\pm$ 0.02}&{\color{olive}95.21\tiny{ $\pm$ 0.02} $\uparrow$ }&95.1\tiny{ $\pm$ 0.04} \\ 
&\task{+ner}&94.98\tiny{ $\pm$ 0.12}&95.09\tiny{ $\pm$ 0.07}&95.05\tiny{ $\pm$ 0.13} \\ 
&\task{+mwe}&{\color{red}94.7\tiny{ $\pm$ 0.16} $\downarrow$ }&{\color{red}94.8\tiny{ $\pm$ 0.08} $\downarrow$ }&{\color{red}94.66\tiny{ $\pm$ 0.07} $\downarrow$ } \\ 
&\task{+sem}&{\color{red}94.77\tiny{ $\pm$ 0.08} $\downarrow$ }&94.82\tiny{ $\pm$ 0.15}&94.93\tiny{ $\pm$ 0.08} \\ 
&\task{+semtr}&{\color{red}94.86\tiny{ $\pm$ 0.02} $\downarrow$ }&{\color{red}94.8\tiny{ $\pm$ 0.09} $\downarrow$ }&94.97\tiny{ $\pm$ 0.09} \\ 
&\task{+supsense}&94.75\tiny{ $\pm$ 0.15}&{\color{red}94.81\tiny{ $\pm$ 0.06} $\downarrow$ }&95.0\tiny{ $\pm$ 0.12} \\ 
&\task{+com}&{\color{red}93.19\tiny{ $\pm$ 0.75} $\downarrow$ }&{\color{red}93.94\tiny{ $\pm$ 0.21} $\downarrow$ }&{\color{red}93.12\tiny{ $\pm$ 0.44} $\downarrow$ } \\ 
&\task{+frame}&{\color{red}94.64\tiny{ $\pm$ 0.06} $\downarrow$ }&{\color{red}94.66\tiny{ $\pm$ 0.05} $\downarrow$ }&{\color{red}94.55\tiny{ $\pm$ 0.06} $\downarrow$ } \\ 
&\task{+hyp}&{\color{red}94.46\tiny{ $\pm$ 0.3} $\downarrow$ }&{\color{red}94.56\tiny{ $\pm$ 0.09} $\downarrow$ }&{\color{red}94.26\tiny{ $\pm$ 0.18} $\downarrow$ } \\  \cline{2-5} 
& Average&94.65&94.77&94.66 \\ \hline 
\parbox[t]{1mm}{\multirow{10}{*}{\rotatebox[origin = c]{90}{All-but-one}}}&All - \task{upos}&{\color{red}94.03\tiny{ $\pm$ 0.13} $\downarrow$ }&{\color{red}94.06\tiny{ $\pm$ 0.09} $\downarrow$ }&{\color{red}94.0\tiny{ $\pm$ 0.26} $\downarrow$ } \\ 
&All - \task{chunk}&{\color{red}94.46\tiny{ $\pm$ 0.09} $\downarrow$ }&{\color{red}94.29\tiny{ $\pm$ 0.07} $\downarrow$ }&{\color{red}94.3\tiny{ $\pm$ 0.12} $\downarrow$ } \\ 
&All - \task{ner}&{\color{red}94.3\tiny{ $\pm$ 0.03} $\downarrow$ }&{\color{red}94.31\tiny{ $\pm$ 0.02} $\downarrow$ }&{\color{red}94.25\tiny{ $\pm$ 0.07} $\downarrow$ } \\ 
&All - \task{mwe}&{\color{red}94.45\tiny{ $\pm$ 0.05} $\downarrow$ }&{\color{red}94.46\tiny{ $\pm$ 0.12} $\downarrow$ }&{\color{red}94.5\tiny{ $\pm$ 0.09} $\downarrow$ } \\ 
&All - \task{sem}&{\color{red}94.34\tiny{ $\pm$ 0.09} $\downarrow$ }&{\color{red}94.41\tiny{ $\pm$ 0.09} $\downarrow$ }&{\color{red}94.32\tiny{ $\pm$ 0.17} $\downarrow$ } \\ 
&All - \task{semtr}&{\color{red}94.35\tiny{ $\pm$ 0.08} $\downarrow$ }&{\color{red}94.28\tiny{ $\pm$ 0.07} $\downarrow$ }&{\color{red}94.25\tiny{ $\pm$ 0.12} $\downarrow$ } \\ 
&All - \task{supsense}&{\color{red}94.54\tiny{ $\pm$ 0.02} $\downarrow$ }&{\color{red}94.4\tiny{ $\pm$ 0.08} $\downarrow$ }&{\color{red}94.27\tiny{ $\pm$ 0.03} $\downarrow$ } \\ 
&All - \task{com}&{\color{red}94.69\tiny{ $\pm$ 0.1} $\downarrow$ }&{\color{red}94.76\tiny{ $\pm$ 0.08} $\downarrow$ }&{\color{red}94.72\tiny{ $\pm$ 0.06} $\downarrow$ } \\ 
&All - \task{frame}&{\color{red}94.57\tiny{ $\pm$ 0.12} $\downarrow$ }&{\color{red}94.6\tiny{ $\pm$ 0.19} $\downarrow$ }&{\color{red}94.39\tiny{ $\pm$ 0.08} $\downarrow$ } \\ 
&All - \task{hyp}&{\color{red}94.53\tiny{ $\pm$ 0.07} $\downarrow$ }&{\color{red}94.45\tiny{ $\pm$ 0.1} $\downarrow$ }&{\color{red}94.45\tiny{ $\pm$ 0.07} $\downarrow$ } \\  \hline 
\multicolumn{2}{c|}{All}&{\color{red}94.31\tiny{ $\pm$ 0.15} $\downarrow$ }&{\color{red}94.42\tiny{ $\pm$ 0.07} $\downarrow$ }&{\color{red}94.3\tiny{ $\pm$ 0.2} $\downarrow$ } \\  \hline 
\multicolumn{2}{c|}{ Oracle } & {\color{blue}95.04 $\pm$ 0.06}&95.21\tiny{ $\pm$ 0.02}&{\color{blue}95.04 $\pm$ 0.06} \\ \hline
\end{tabular}
\caption{\small F1 score tested on the task \task{xpos} in different training scenarios}\label{tMultiTaskxposuni}}}
\parbox{.49\linewidth}{
\centering
\scriptsize{
\begin{tabular}{c|c|c|c|c}
\multicolumn{2}{c}{Trained with} & \multicolumn{3}{|c}{Tested on \task{chunk}} \\ \cline{3-5}
 \multicolumn{2}{c|}{} & \multidec & \tedec & \teenc \\ \hline
\multicolumn{2}{c}{ \task{chunk} only } & \multicolumn{3}{|c}{93.49\tiny{ $\pm$ 0.01}}\\ \hline
\parbox[t]{1mm}{\multirow{11}{*}{\rotatebox[origin = c]{90}{Pairwise}}}&\task{+upos}&{\color{olive}94.18\tiny{ $\pm$ 0.02} $\uparrow$ }&{\color{olive}94.02\tiny{ $\pm$ 0.08} $\uparrow$ }&{\color{olive}94.0\tiny{ $\pm$ 0.15} $\uparrow$ } \\ 
&\task{+xpos}&{\color{olive}93.97\tiny{ $\pm$ 0.16} $\uparrow$ }&{\color{olive}94.18\tiny{ $\pm$ 0.01} $\uparrow$ }&{\color{olive}93.98\tiny{ $\pm$ 0.13} $\uparrow$ } \\ 
&\task{+ner}&93.47\tiny{ $\pm$ 0.1}&{\color{olive}93.64\tiny{ $\pm$ 0.03} $\uparrow$ }&93.54\tiny{ $\pm$ 0.1} \\ 
&\task{+mwe}&93.54\tiny{ $\pm$ 0.13}&93.59\tiny{ $\pm$ 0.2}&93.33\tiny{ $\pm$ 0.2} \\ 
&\task{+sem}&{\color{olive}93.63\tiny{ $\pm$ 0.02} $\uparrow$ }&93.45\tiny{ $\pm$ 0.07}&93.52\tiny{ $\pm$ 0.13} \\ 
&\task{+semtr}&93.61\tiny{ $\pm$ 0.07}&93.47\tiny{ $\pm$ 0.03}&93.45\tiny{ $\pm$ 0.07} \\ 
&\task{+supsense}&93.2\tiny{ $\pm$ 0.21}&93.25\tiny{ $\pm$ 0.15}&{\color{red}93.13\tiny{ $\pm$ 0.13} $\downarrow$ } \\ 
&\task{+com}&{\color{red}91.94\tiny{ $\pm$ 0.4} $\downarrow$ }&{\color{red}92.29\tiny{ $\pm$ 0.27} $\downarrow$ }&{\color{red}91.86\tiny{ $\pm$ 0.09} $\downarrow$ } \\ 
&\task{+frame}&{\color{red}93.22\tiny{ $\pm$ 0.16} $\downarrow$ }&{\color{red}93.23\tiny{ $\pm$ 0.04} $\downarrow$ }&93.29\tiny{ $\pm$ 0.13} \\ 
&\task{+hyp}&{\color{red}92.96\tiny{ $\pm$ 0.08} $\downarrow$ }&{\color{red}92.86\tiny{ $\pm$ 0.08} $\downarrow$ }&{\color{red}93.13\tiny{ $\pm$ 0.04} $\downarrow$ } \\  \cline{2-5} 
& Average&93.37&93.4&93.32 \\ \hline 
\parbox[t]{1mm}{\multirow{10}{*}{\rotatebox[origin = c]{90}{All-but-one}}}&All - \task{upos}&93.59\tiny{ $\pm$ 0.13}&93.44\tiny{ $\pm$ 0.17}&93.36\tiny{ $\pm$ 0.17} \\ 
&All - \task{xpos}&93.57\tiny{ $\pm$ 0.19}&{\color{olive}93.6\tiny{ $\pm$ 0.05} $\uparrow$ }&93.29\tiny{ $\pm$ 0.21} \\ 
&All - \task{ner}&93.59\tiny{ $\pm$ 0.09}&93.69\tiny{ $\pm$ 0.14}&93.5\tiny{ $\pm$ 0.23} \\ 
&All - \task{mwe}&{\color{olive}93.71\tiny{ $\pm$ 0.11} $\uparrow$ }&{\color{olive}93.72\tiny{ $\pm$ 0.13} $\uparrow$ }&{\color{olive}93.63\tiny{ $\pm$ 0.04} $\uparrow$ } \\ 
&All - \task{sem}&93.63\tiny{ $\pm$ 0.08}&93.6\tiny{ $\pm$ 0.11}&93.45\tiny{ $\pm$ 0.13} \\ 
&All - \task{semtr}&93.58\tiny{ $\pm$ 0.08}&93.56\tiny{ $\pm$ 0.14}&93.54\tiny{ $\pm$ 0.06} \\ 
&All - \task{supsense}&{\color{olive}93.67\tiny{ $\pm$ 0.08} $\uparrow$ }&93.67\tiny{ $\pm$ 0.12}&93.56\tiny{ $\pm$ 0.12} \\ 
&All - \task{com}&93.67\tiny{ $\pm$ 0.12}&{\color{olive}93.79\tiny{ $\pm$ 0.14} $\uparrow$ }&{\color{olive}93.82\tiny{ $\pm$ 0.05} $\uparrow$ } \\ 
&All - \task{frame}&{\color{olive}93.7\tiny{ $\pm$ 0.09} $\uparrow$ }&93.64\tiny{ $\pm$ 0.11}&93.51\tiny{ $\pm$ 0.06} \\ 
&All - \task{hyp}&{\color{olive}93.78\tiny{ $\pm$ 0.12} $\uparrow$ }&{\color{olive}93.69\tiny{ $\pm$ 0.05} $\uparrow$ }&93.59\tiny{ $\pm$ 0.07} \\  \hline 
\multicolumn{2}{c|}{All}&93.44\tiny{ $\pm$ 0.09}&93.64\tiny{ $\pm$ 0.21}&{\color{olive}93.7\tiny{ $\pm$ 0.06} $\uparrow$ } \\  \hline 
\multicolumn{2}{c|}{ Oracle } & 94.01\tiny{ $\pm$ 0.13}&94.07\tiny{ $\pm$ 0.25}&93.93\tiny{ $\pm$ 0.16} \\ \hline
\end{tabular}
\caption{\small F1 score tested on the task \task{chunk} in different training scenarios}\label{tMultiTaskchunkconll02}}}
\parbox{.49\linewidth}{
\centering
\scriptsize{
\begin{tabular}{c|c|c|c|c}
\multicolumn{2}{c}{Trained with} & \multicolumn{3}{|c}{Tested on \task{ner}} \\ \cline{3-5}
 \multicolumn{2}{c|}{} & \multidec & \tedec & \teenc \\ \hline
\multicolumn{2}{c}{ \task{ner} only } & \multicolumn{3}{|c}{88.24\tiny{ $\pm$ 0.09}}\\ \hline
\parbox[t]{1mm}{\multirow{11}{*}{\rotatebox[origin = c]{90}{Pairwise}}}&\task{+upos}&87.68\tiny{ $\pm$ 0.41}&87.99\tiny{ $\pm$ 0.21}&{\color{red}87.43\tiny{ $\pm$ 0.11} $\downarrow$ } \\ 
&\task{+xpos}&{\color{red}87.61\tiny{ $\pm$ 0.27} $\downarrow$ }&{\color{red}87.65\tiny{ $\pm$ 0.14} $\downarrow$ }&{\color{red}87.71\tiny{ $\pm$ 0.08} $\downarrow$ } \\ 
&\task{+chunk}&87.96\tiny{ $\pm$ 0.19}&88.11\tiny{ $\pm$ 0.21}&88.07\tiny{ $\pm$ 0.16} \\ 
&\task{+mwe}&88.15\tiny{ $\pm$ 0.23}&87.99\tiny{ $\pm$ 0.15}&88.02\tiny{ $\pm$ 0.36} \\ 
&\task{+sem}&{\color{red}87.35\tiny{ $\pm$ 0.16} $\downarrow$ }&{\color{red}87.27\tiny{ $\pm$ 0.36} $\downarrow$ }&{\color{red}87.49\tiny{ $\pm$ 0.25} $\downarrow$ } \\ 
&\task{+semtr}&{\color{red}87.34\tiny{ $\pm$ 0.27} $\downarrow$ }&87.75\tiny{ $\pm$ 0.38}&{\color{red}87.29\tiny{ $\pm$ 0.17} $\downarrow$ } \\ 
&\task{+supsense}&87.9\tiny{ $\pm$ 0.24}&87.94\tiny{ $\pm$ 0.33}&87.92\tiny{ $\pm$ 0.16} \\ 
&\task{+com}&{\color{red}86.62\tiny{ $\pm$ 0.72} $\downarrow$ }&{\color{red}86.59\tiny{ $\pm$ 0.31} $\downarrow$ }&{\color{red}86.75\tiny{ $\pm$ 0.45} $\downarrow$ } \\ 
&\task{+frame}&88.15\tiny{ $\pm$ 0.35}&88.02\tiny{ $\pm$ 0.17}&87.99\tiny{ $\pm$ 0.32} \\ 
&\task{+hyp}&87.98\tiny{ $\pm$ 0.21}&87.91\tiny{ $\pm$ 0.4}&87.82\tiny{ $\pm$ 0.31} \\  \cline{2-5} 
& Average&87.67&87.72&87.65 \\ \hline 
\parbox[t]{1mm}{\multirow{10}{*}{\rotatebox[origin = c]{90}{All-but-one}}}&All - \task{upos}&{\color{red}86.03\tiny{ $\pm$ 0.53} $\downarrow$ }&{\color{red}86.47\tiny{ $\pm$ 0.14} $\downarrow$ }&{\color{red}85.98\tiny{ $\pm$ 0.29} $\downarrow$ } \\ 
&All - \task{xpos}&{\color{red}86.04\tiny{ $\pm$ 0.15} $\downarrow$ }&{\color{red}86.68\tiny{ $\pm$ 0.27} $\downarrow$ }&{\color{red}85.8\tiny{ $\pm$ 0.27} $\downarrow$ } \\ 
&All - \task{chunk}&{\color{red}86.05\tiny{ $\pm$ 0.1} $\downarrow$ }&{\color{red}86.08\tiny{ $\pm$ 0.49} $\downarrow$ }&{\color{red}85.73\tiny{ $\pm$ 0.2} $\downarrow$ } \\ 
&All - \task{mwe}&{\color{red}86.21\tiny{ $\pm$ 0.27} $\downarrow$ }&{\color{red}86.21\tiny{ $\pm$ 0.19} $\downarrow$ }&{\color{red}86.1\tiny{ $\pm$ 0.37} $\downarrow$ } \\ 
&All - \task{sem}&{\color{red}85.81\tiny{ $\pm$ 0.32} $\downarrow$ }&{\color{red}85.97\tiny{ $\pm$ 0.14} $\downarrow$ }&{\color{red}85.58\tiny{ $\pm$ 0.04} $\downarrow$ } \\ 
&All - \task{semtr}&{\color{red}86.11\tiny{ $\pm$ 0.28} $\downarrow$ }&{\color{red}86.23\tiny{ $\pm$ 0.23} $\downarrow$ }&{\color{red}86.02\tiny{ $\pm$ 0.39} $\downarrow$ } \\ 
&All - \task{supsense}&{\color{red}86.43\tiny{ $\pm$ 0.12} $\downarrow$ }&{\color{red}86.49\tiny{ $\pm$ 0.17} $\downarrow$ }&{\color{red}86.04\tiny{ $\pm$ 0.14} $\downarrow$ } \\ 
&All - \task{com}&{\color{red}86.6\tiny{ $\pm$ 0.79} $\downarrow$ }&{\color{red}86.25\tiny{ $\pm$ 0.06} $\downarrow$ }&{\color{red}86.23\tiny{ $\pm$ 0.33} $\downarrow$ } \\ 
&All - \task{frame}&{\color{red}85.9\tiny{ $\pm$ 0.29} $\downarrow$ }&{\color{red}86.68\tiny{ $\pm$ 0.15} $\downarrow$ }&{\color{red}85.99\tiny{ $\pm$ 0.3} $\downarrow$ } \\ 
&All - \task{hyp}&{\color{red}86.31\tiny{ $\pm$ 0.18} $\downarrow$ }&{\color{red}86.86\tiny{ $\pm$ 0.25} $\downarrow$ }&{\color{red}86.1\tiny{ $\pm$ 0.56} $\downarrow$ } \\  \hline 
\multicolumn{2}{c|}{All}&{\color{red}86.38\tiny{ $\pm$ 0.12} $\downarrow$ }&{\color{red}86.8\tiny{ $\pm$ 0.08} $\downarrow$ }&{\color{red}86.01\tiny{ $\pm$ 0.4} $\downarrow$ } \\  \hline 
\multicolumn{2}{c|}{ Oracle } & {\color{blue}88.24 $\pm$ 0.09}&{\color{blue}88.24 $\pm$ 0.09}&{\color{blue}88.24 $\pm$ 0.09} \\ \hline
\end{tabular}
\caption{\small F1 score tested on the task \task{ner} in different training scenarios}\label{tMultiTasknerconll03}}}
\end{table*}

\begin{table*}[t]
\parbox{.49\linewidth}{
\centering
\scriptsize{
\begin{tabular}{c|c|c|c|c}
\multicolumn{2}{c}{Trained with} & \multicolumn{3}{|c}{Tested on \task{mwe}} \\ \cline{3-5}
 \multicolumn{2}{c|}{} & \multidec & \tedec & \teenc \\ \hline
\multicolumn{2}{c}{ \task{mwe} only } & \multicolumn{3}{|c}{53.07\tiny{ $\pm$ 0.12}}\\ \hline
\parbox[t]{1mm}{\multirow{11}{*}{\rotatebox[origin = c]{90}{Pairwise}}}&\task{+upos}&{\color{olive}59.99\tiny{ $\pm$ 0.36} $\uparrow$ }&{\color{olive}60.28\tiny{ $\pm$ 0.24} $\uparrow$ }&{\color{olive}57.61\tiny{ $\pm$ 0.2} $\uparrow$ } \\ 
&\task{+xpos}&{\color{olive}58.87\tiny{ $\pm$ 0.78} $\uparrow$ }&{\color{olive}60.32\tiny{ $\pm$ 0.3} $\uparrow$ }&{\color{olive}58.26\tiny{ $\pm$ 0.25} $\uparrow$ } \\ 
&\task{+chunk}&{\color{olive}59.18\tiny{ $\pm$ 0.03} $\uparrow$ }&{\color{olive}57.61\tiny{ $\pm$ 1.53} $\uparrow$ }&{\color{olive}58.06\tiny{ $\pm$ 0.88} $\uparrow$ } \\ 
&\task{+ner}&{\color{olive}55.4\tiny{ $\pm$ 0.52} $\uparrow$ }&{\color{olive}55.17\tiny{ $\pm$ 0.44} $\uparrow$ }&53.4\tiny{ $\pm$ 0.98} \\ 
&\task{+sem}&{\color{olive}60.16\tiny{ $\pm$ 1.23} $\uparrow$ }&{\color{olive}58.21\tiny{ $\pm$ 0.09} $\uparrow$ }&{\color{olive}58.62\tiny{ $\pm$ 0.61} $\uparrow$ } \\ 
&\task{+semtr}&{\color{olive}58.84\tiny{ $\pm$ 1.45} $\uparrow$ }&{\color{olive}58.55\tiny{ $\pm$ 0.28} $\uparrow$ }&{\color{olive}58.31\tiny{ $\pm$ 2.24} $\uparrow$ } \\ 
&\task{+supsense}&{\color{olive}58.81\tiny{ $\pm$ 1.01} $\uparrow$ }&{\color{olive}58.75\tiny{ $\pm$ 0.33} $\uparrow$ }&{\color{olive}58.05\tiny{ $\pm$ 0.72} $\uparrow$ } \\ 
&\task{+com}&53.89\tiny{ $\pm$ 1.41}&51.72\tiny{ $\pm$ 1.01}&51.71\tiny{ $\pm$ 1.05} \\ 
&\task{+frame}&53.88\tiny{ $\pm$ 0.76}&53.05\tiny{ $\pm$ 1.32}&53.3\tiny{ $\pm$ 1.15} \\ 
&\task{+hyp}&53.08\tiny{ $\pm$ 1.72}&52.98\tiny{ $\pm$ 1.66}&52.59\tiny{ $\pm$ 1.98} \\  \cline{2-5} 
& Average&57.21&56.67&55.99 \\ \hline 
\parbox[t]{1mm}{\multirow{10}{*}{\rotatebox[origin = c]{90}{All-but-one}}}&All - \task{upos}&{\color{olive}61.28\tiny{ $\pm$ 0.78} $\uparrow$ }&{\color{olive}60.48\tiny{ $\pm$ 0.93} $\uparrow$ }&{\color{olive}59.58\tiny{ $\pm$ 1.14} $\uparrow$ } \\ 
&All - \task{xpos}&{\color{olive}61.91\tiny{ $\pm$ 1.56} $\uparrow$ }&{\color{olive}60.09\tiny{ $\pm$ 0.9} $\uparrow$ }&{\color{olive}59.81\tiny{ $\pm$ 0.83} $\uparrow$ } \\ 
&All - \task{chunk}&{\color{olive}61.01\tiny{ $\pm$ 1.61} $\uparrow$ }&{\color{olive}60.6\tiny{ $\pm$ 1.52} $\uparrow$ }&{\color{olive}61.58\tiny{ $\pm$ 1.05} $\uparrow$ } \\ 
&All - \task{ner}&{\color{olive}62.69\tiny{ $\pm$ 0.26} $\uparrow$ }&{\color{olive}60.48\tiny{ $\pm$ 0.15} $\uparrow$ }&{\color{olive}59.05\tiny{ $\pm$ 0.4} $\uparrow$ } \\ 
&All - \task{sem}&{\color{olive}61.17\tiny{ $\pm$ 0.86} $\uparrow$ }&{\color{olive}59.94\tiny{ $\pm$ 0.85} $\uparrow$ }&{\color{olive}59.47\tiny{ $\pm$ 0.04} $\uparrow$ } \\ 
&All - \task{semtr}&{\color{olive}63.04\tiny{ $\pm$ 0.85} $\uparrow$ }&{\color{olive}61.23\tiny{ $\pm$ 2.05} $\uparrow$ }&{\color{olive}60.59\tiny{ $\pm$ 0.59} $\uparrow$ } \\ 
&All - \task{supsense}&{\color{olive}60.51\tiny{ $\pm$ 0.25} $\uparrow$ }&{\color{olive}59.11\tiny{ $\pm$ 2.02} $\uparrow$ }&{\color{olive}59.25\tiny{ $\pm$ 0.74} $\uparrow$ } \\ 
&All - \task{com}&{\color{olive}61.95\tiny{ $\pm$ 0.97} $\uparrow$ }&{\color{olive}62.02\tiny{ $\pm$ 1.73} $\uparrow$ }&{\color{olive}60.63\tiny{ $\pm$ 0.73} $\uparrow$ } \\ 
&All - \task{frame}&{\color{olive}62.62\tiny{ $\pm$ 0.85} $\uparrow$ }&{\color{olive}60.52\tiny{ $\pm$ 0.47} $\uparrow$ }&{\color{olive}61.21\tiny{ $\pm$ 0.99} $\uparrow$ } \\ 
&All - \task{hyp}&{\color{olive}62.04\tiny{ $\pm$ 0.6} $\uparrow$ }&{\color{olive}61.07\tiny{ $\pm$ 0.51} $\uparrow$ }&{\color{olive}61.09\tiny{ $\pm$ 1.06} $\uparrow$ } \\  \hline 
\multicolumn{2}{c|}{All}&{\color{olive}61.43\tiny{ $\pm$ 1.94} $\uparrow$ }&{\color{olive}61.97\tiny{ $\pm$ 0.5} $\uparrow$ }&{\color{olive}59.57\tiny{ $\pm$ 0.64} $\uparrow$ } \\  \hline 
\multicolumn{2}{c|}{ Oracle } & 62.76\tiny{ $\pm$ 0.63}&61.74\tiny{ $\pm$ 1.49}&61.92\tiny{ $\pm$ 0.66} \\ \hline
\end{tabular}
\caption{\small F1 score tested on the task \task{mwe} in different training scenarios}\label{tMultiTaskmwestreusle}}}
\parbox{.49\linewidth}{
\centering
\scriptsize{
\begin{tabular}{c|c|c|c|c}
\multicolumn{2}{c}{Trained with} & \multicolumn{3}{|c}{Tested on \task{sem}} \\ \cline{3-5}
 \multicolumn{2}{c|}{} & \multidec & \tedec & \teenc \\ \hline
\multicolumn{2}{c}{ \task{sem} only } & \multicolumn{3}{|c}{72.77\tiny{ $\pm$ 0.04}}\\ \hline
\parbox[t]{1mm}{\multirow{11}{*}{\rotatebox[origin = c]{90}{Pairwise}}}&\task{+upos}&{\color{olive}73.23\tiny{ $\pm$ 0.06} $\uparrow$ }&{\color{olive}73.17\tiny{ $\pm$ 0.08} $\uparrow$ }&{\color{olive}73.11\tiny{ $\pm$ 0.01} $\uparrow$ } \\ 
&\task{+xpos}&{\color{olive}73.34\tiny{ $\pm$ 0.12} $\uparrow$ }&{\color{olive}73.21\tiny{ $\pm$ 0.04} $\uparrow$ }&73.04\tiny{ $\pm$ 0.21} \\ 
&\task{+chunk}&{\color{olive}73.16\tiny{ $\pm$ 0.05} $\uparrow$ }&{\color{olive}73.02\tiny{ $\pm$ 0.05} $\uparrow$ }&{\color{olive}73.13\tiny{ $\pm$ 0.07} $\uparrow$ } \\ 
&\task{+ner}&72.88\tiny{ $\pm$ 0.08}&72.77\tiny{ $\pm$ 0.19}&72.91\tiny{ $\pm$ 0.08} \\ 
&\task{+mwe}&72.75\tiny{ $\pm$ 0.09}&72.66\tiny{ $\pm$ 0.18}&72.83\tiny{ $\pm$ 0.07} \\ 
&\task{+semtr}&{\color{red}72.5\tiny{ $\pm$ 0.07} $\downarrow$ }&{\color{red}72.5\tiny{ $\pm$ 0.05} $\downarrow$ }&{\color{red}72.17\tiny{ $\pm$ 0.06} $\downarrow$ } \\ 
&\task{+supsense}&72.81\tiny{ $\pm$ 0.04}&72.71\tiny{ $\pm$ 0.03}&{\color{olive}73.09\tiny{ $\pm$ 0.08} $\uparrow$ } \\ 
&\task{+com}&{\color{red}70.39\tiny{ $\pm$ 0.46} $\downarrow$ }&{\color{red}70.37\tiny{ $\pm$ 0.28} $\downarrow$ }&{\color{red}70.18\tiny{ $\pm$ 0.54} $\downarrow$ } \\ 
&\task{+frame}&72.76\tiny{ $\pm$ 0.16}&{\color{red}72.26\tiny{ $\pm$ 0.21} $\downarrow$ }&72.49\tiny{ $\pm$ 0.23} \\ 
&\task{+hyp}&{\color{red}72.47\tiny{ $\pm$ 0.02} $\downarrow$ }&{\color{red}72.15\tiny{ $\pm$ 0.1} $\downarrow$ }&71.95\tiny{ $\pm$ 1.22} \\  \cline{2-5} 
& Average&72.63&72.48&72.49 \\ \hline 
\parbox[t]{1mm}{\multirow{10}{*}{\rotatebox[origin = c]{90}{All-but-one}}}&All - \task{upos}&{\color{red}70.87\tiny{ $\pm$ 0.19} $\downarrow$ }&{\color{red}71.08\tiny{ $\pm$ 0.19} $\downarrow$ }&{\color{red}70.68\tiny{ $\pm$ 0.76} $\downarrow$ } \\ 
&All - \task{xpos}&{\color{red}71.12\tiny{ $\pm$ 0.1} $\downarrow$ }&{\color{red}70.98\tiny{ $\pm$ 0.24} $\downarrow$ }&{\color{red}70.57\tiny{ $\pm$ 0.13} $\downarrow$ } \\ 
&All - \task{chunk}&{\color{red}71.07\tiny{ $\pm$ 0.27} $\downarrow$ }&{\color{red}70.39\tiny{ $\pm$ 0.39} $\downarrow$ }&{\color{red}70.78\tiny{ $\pm$ 0.35} $\downarrow$ } \\ 
&All - \task{ner}&{\color{red}70.82\tiny{ $\pm$ 0.41} $\downarrow$ }&{\color{red}70.64\tiny{ $\pm$ 0.15} $\downarrow$ }&{\color{red}70.58\tiny{ $\pm$ 0.03} $\downarrow$ } \\ 
&All - \task{mwe}&{\color{red}71.01\tiny{ $\pm$ 0.14} $\downarrow$ }&{\color{red}71.11\tiny{ $\pm$ 0.17} $\downarrow$ }&{\color{red}71.12\tiny{ $\pm$ 0.29} $\downarrow$ } \\ 
&All - \task{semtr}&{\color{red}69.72\tiny{ $\pm$ 0.27} $\downarrow$ }&{\color{red}69.62\tiny{ $\pm$ 0.37} $\downarrow$ }&{\color{red}69.86\tiny{ $\pm$ 0.36} $\downarrow$ } \\ 
&All - \task{supsense}&{\color{red}71.22\tiny{ $\pm$ 0.29} $\downarrow$ }&{\color{red}71.02\tiny{ $\pm$ 0.16} $\downarrow$ }&{\color{red}70.53\tiny{ $\pm$ 0.19} $\downarrow$ } \\ 
&All - \task{com}&{\color{red}72.38\tiny{ $\pm$ 0.08} $\downarrow$ }&{\color{red}72.32\tiny{ $\pm$ 0.23} $\downarrow$ }&{\color{red}72.38\tiny{ $\pm$ 0.17} $\downarrow$ } \\ 
&All - \task{frame}&{\color{red}71.48\tiny{ $\pm$ 0.51} $\downarrow$ }&{\color{red}71.11\tiny{ $\pm$ 0.16} $\downarrow$ }&{\color{red}70.78\tiny{ $\pm$ 0.44} $\downarrow$ } \\ 
&All - \task{hyp}&{\color{red}71.22\tiny{ $\pm$ 0.25} $\downarrow$ }&{\color{red}71.22\tiny{ $\pm$ 0.33} $\downarrow$ }&{\color{red}71.03\tiny{ $\pm$ 0.07} $\downarrow$ } \\  \hline 
\multicolumn{2}{c|}{All}&{\color{red}71.53\tiny{ $\pm$ 0.28} $\downarrow$ }&{\color{red}71.72\tiny{ $\pm$ 0.21} $\downarrow$ }&{\color{red}71.58\tiny{ $\pm$ 0.24} $\downarrow$ } \\  \hline 
\multicolumn{2}{c|}{ Oracle } & 73.32\tiny{ $\pm$ 0.04}&73.1\tiny{ $\pm$ 0.03}&73.14\tiny{ $\pm$ 0.06} \\ \hline
\end{tabular}
\caption{\small F1 score tested on the task \task{sem} in different training scenarios}\label{tMultiTasksemsemcor}}}
\parbox{.49\linewidth}{
\centering
\scriptsize{
\begin{tabular}{c|c|c|c|c}
\multicolumn{2}{c}{Trained with} & \multicolumn{3}{|c}{Tested on \task{semtr}} \\ \cline{3-5}
 \multicolumn{2}{c|}{} & \multidec & \tedec & \teenc \\ \hline
\multicolumn{2}{c}{ \task{semtr} only } & \multicolumn{3}{|c}{74.02\tiny{ $\pm$ 0.04}}\\ \hline
\parbox[t]{1mm}{\multirow{11}{*}{\rotatebox[origin = c]{90}{Pairwise}}}&\task{+upos}&{\color{olive}74.93\tiny{ $\pm$ 0.09} $\uparrow$ }&{\color{olive}74.87\tiny{ $\pm$ 0.1} $\uparrow$ }&{\color{olive}74.85\tiny{ $\pm$ 0.05} $\uparrow$ } \\ 
&\task{+xpos}&{\color{olive}74.91\tiny{ $\pm$ 0.06} $\uparrow$ }&{\color{olive}74.84\tiny{ $\pm$ 0.21} $\uparrow$ }&{\color{olive}74.66\tiny{ $\pm$ 0.2} $\uparrow$ } \\ 
&\task{+chunk}&{\color{olive}74.79\tiny{ $\pm$ 0.13} $\uparrow$ }&{\color{olive}74.73\tiny{ $\pm$ 0.12} $\uparrow$ }&{\color{olive}74.77\tiny{ $\pm$ 0.13} $\uparrow$ } \\ 
&\task{+ner}&{\color{olive}74.34\tiny{ $\pm$ 0.08} $\uparrow$ }&74.01\tiny{ $\pm$ 0.05}&74.04\tiny{ $\pm$ 0.07} \\ 
&\task{+mwe}&{\color{olive}74.51\tiny{ $\pm$ 0.18} $\uparrow$ }&{\color{olive}74.63\tiny{ $\pm$ 0.28} $\uparrow$ }&{\color{olive}74.66\tiny{ $\pm$ 0.21} $\uparrow$ } \\ 
&\task{+sem}&{\color{olive}74.73\tiny{ $\pm$ 0.1} $\uparrow$ }&{\color{olive}74.72\tiny{ $\pm$ 0.14} $\uparrow$ }&{\color{olive}74.41\tiny{ $\pm$ 0.01} $\uparrow$ } \\ 
&\task{+supsense}&{\color{olive}74.61\tiny{ $\pm$ 0.24} $\uparrow$ }&{\color{olive}74.52\tiny{ $\pm$ 0.05} $\uparrow$ }&{\color{olive}74.94\tiny{ $\pm$ 0.22} $\uparrow$ } \\ 
&\task{+com}&72.6\tiny{ $\pm$ 0.95}&{\color{red}71.76\tiny{ $\pm$ 0.88} $\downarrow$ }&{\color{red}71.35\tiny{ $\pm$ 0.95} $\downarrow$ } \\ 
&\task{+frame}&74.18\tiny{ $\pm$ 0.19}&74.21\tiny{ $\pm$ 0.37}&{\color{olive}74.63\tiny{ $\pm$ 0.11} $\uparrow$ } \\ 
&\task{+hyp}&74.23\tiny{ $\pm$ 0.27}&74.19\tiny{ $\pm$ 0.45}&74.14\tiny{ $\pm$ 0.23} \\  \cline{2-5} 
& Average&74.38&74.25&74.24 \\ \hline 
\parbox[t]{1mm}{\multirow{10}{*}{\rotatebox[origin = c]{90}{All-but-one}}}&All - \task{upos}&73.54\tiny{ $\pm$ 0.54}&73.79\tiny{ $\pm$ 0.46}&73.66\tiny{ $\pm$ 0.97} \\ 
&All - \task{xpos}&74.03\tiny{ $\pm$ 0.11}&73.78\tiny{ $\pm$ 0.28}&{\color{red}73.64\tiny{ $\pm$ 0.07} $\downarrow$ } \\ 
&All - \task{chunk}&73.97\tiny{ $\pm$ 0.22}&{\color{red}73.36\tiny{ $\pm$ 0.05} $\downarrow$ }&73.65\tiny{ $\pm$ 0.39} \\ 
&All - \task{ner}&73.51\tiny{ $\pm$ 0.35}&{\color{red}73.59\tiny{ $\pm$ 0.19} $\downarrow$ }&{\color{red}73.4\tiny{ $\pm$ 0.19} $\downarrow$ } \\ 
&All - \task{mwe}&{\color{red}73.61\tiny{ $\pm$ 0.2} $\downarrow$ }&74.04\tiny{ $\pm$ 0.18}&73.75\tiny{ $\pm$ 0.24} \\ 
&All - \task{sem}&{\color{red}71.97\tiny{ $\pm$ 0.3} $\downarrow$ }&{\color{red}72.26\tiny{ $\pm$ 0.28} $\downarrow$ }&{\color{red}72.21\tiny{ $\pm$ 0.48} $\downarrow$ } \\ 
&All - \task{supsense}&73.86\tiny{ $\pm$ 0.09}&73.76\tiny{ $\pm$ 0.19}&{\color{red}73.27\tiny{ $\pm$ 0.2} $\downarrow$ } \\ 
&All - \task{com}&{\color{olive}74.75\tiny{ $\pm$ 0.22} $\uparrow$ }&{\color{olive}74.92\tiny{ $\pm$ 0.1} $\uparrow$ }&{\color{olive}75.06\tiny{ $\pm$ 0.12} $\uparrow$ } \\ 
&All - \task{frame}&74.24\tiny{ $\pm$ 0.37}&73.9\tiny{ $\pm$ 0.29}&73.69\tiny{ $\pm$ 0.32} \\ 
&All - \task{hyp}&74.02\tiny{ $\pm$ 0.12}&74.04\tiny{ $\pm$ 0.17}&74.09\tiny{ $\pm$ 0.21} \\  \hline 
\multicolumn{2}{c|}{All}&{\color{olive}74.26\tiny{ $\pm$ 0.1} $\uparrow$ }&{\color{olive}74.36\tiny{ $\pm$ 0.03} $\uparrow$ }&74.35\tiny{ $\pm$ 0.29} \\  \hline 
\multicolumn{2}{c|}{ Oracle } & 75.23\tiny{ $\pm$ 0.06}&75.24\tiny{ $\pm$ 0.13}&75.09\tiny{ $\pm$ 0.02} \\ \hline
\end{tabular}
\caption{\small F1 score tested on the task \task{semtr} in different training scenarios}\label{tMultiTasksemtrsemcor}}}
\parbox{.49\linewidth}{
\centering
\scriptsize{
\begin{tabular}{c|c|c|c|c}
\multicolumn{2}{c}{Trained with} & \multicolumn{3}{|c}{Tested on \task{supsense}} \\ \cline{3-5}
 \multicolumn{2}{c|}{} & \multidec & \tedec & \teenc \\ \hline
\multicolumn{2}{c}{ \task{supsense} only } & \multicolumn{3}{|c}{66.81\tiny{ $\pm$ 0.22}}\\ \hline
\parbox[t]{1mm}{\multirow{11}{*}{\rotatebox[origin = c]{90}{Pairwise}}}&\task{+upos}&{\color{olive}68.25\tiny{ $\pm$ 0.42} $\uparrow$ }&{\color{olive}67.8\tiny{ $\pm$ 0.29} $\uparrow$ }&{\color{olive}67.76\tiny{ $\pm$ 0.14} $\uparrow$ } \\ 
&\task{+xpos}&{\color{olive}67.78\tiny{ $\pm$ 0.4} $\uparrow$ }&{\color{olive}68.3\tiny{ $\pm$ 0.71} $\uparrow$ }&{\color{olive}67.77\tiny{ $\pm$ 0.15} $\uparrow$ } \\ 
&\task{+chunk}&{\color{olive}67.39\tiny{ $\pm$ 0.15} $\uparrow$ }&67.29\tiny{ $\pm$ 0.33}&67.36\tiny{ $\pm$ 0.29} \\ 
&\task{+ner}&{\color{olive}68.06\tiny{ $\pm$ 0.16} $\uparrow$ }&67.25\tiny{ $\pm$ 0.21}&{\color{olive}67.57\tiny{ $\pm$ 0.27} $\uparrow$ } \\ 
&\task{+mwe}&66.88\tiny{ $\pm$ 0.14}&66.88\tiny{ $\pm$ 0.24}&66.26\tiny{ $\pm$ 0.9} \\ 
&\task{+sem}&{\color{olive}68.29\tiny{ $\pm$ 0.21} $\uparrow$ }&{\color{olive}68.46\tiny{ $\pm$ 0.38} $\uparrow$ }&{\color{olive}68.1\tiny{ $\pm$ 0.59} $\uparrow$ } \\ 
&\task{+semtr}&{\color{olive}68.6\tiny{ $\pm$ 0.81} $\uparrow$ }&{\color{olive}68.18\tiny{ $\pm$ 0.39} $\uparrow$ }&67.64\tiny{ $\pm$ 0.92} \\ 
&\task{+com}&{\color{red}65.57\tiny{ $\pm$ 0.17} $\downarrow$ }&{\color{red}64.98\tiny{ $\pm$ 0.34} $\downarrow$ }&{\color{red}65.55\tiny{ $\pm$ 0.18} $\downarrow$ } \\ 
&\task{+frame}&66.59\tiny{ $\pm$ 0.07}&{\color{red}66.2\tiny{ $\pm$ 0.16} $\downarrow$ }&66.75\tiny{ $\pm$ 0.22} \\ 
&\task{+hyp}&66.47\tiny{ $\pm$ 0.24}&66.52\tiny{ $\pm$ 0.59}&66.16\tiny{ $\pm$ 0.43} \\  \cline{2-5} 
& Average&67.39&67.19&67.09 \\ \hline 
\parbox[t]{1mm}{\multirow{10}{*}{\rotatebox[origin = c]{90}{All-but-one}}}&All - \task{upos}&{\color{olive}68.27\tiny{ $\pm$ 0.33} $\uparrow$ }&{\color{olive}68.1\tiny{ $\pm$ 0.28} $\uparrow$ }&{\color{olive}68.19\tiny{ $\pm$ 0.55} $\uparrow$ } \\ 
&All - \task{xpos}&{\color{olive}67.99\tiny{ $\pm$ 0.5} $\uparrow$ }&67.9\tiny{ $\pm$ 0.54}&{\color{olive}68.47\tiny{ $\pm$ 0.18} $\uparrow$ } \\ 
&All - \task{chunk}&{\color{olive}68.26\tiny{ $\pm$ 0.48} $\uparrow$ }&{\color{olive}68.07\tiny{ $\pm$ 0.28} $\uparrow$ }&{\color{olive}67.87\tiny{ $\pm$ 0.32} $\uparrow$ } \\ 
&All - \task{ner}&{\color{olive}68.16\tiny{ $\pm$ 0.26} $\uparrow$ }&67.51\tiny{ $\pm$ 0.4}&{\color{olive}67.95\tiny{ $\pm$ 0.24} $\uparrow$ } \\ 
&All - \task{mwe}&{\color{olive}68.18\tiny{ $\pm$ 0.62} $\uparrow$ }&67.38\tiny{ $\pm$ 0.22}&{\color{olive}69.0\tiny{ $\pm$ 0.45} $\uparrow$ } \\ 
&All - \task{sem}&67.36\tiny{ $\pm$ 0.42}&67.35\tiny{ $\pm$ 0.18}&{\color{olive}67.77\tiny{ $\pm$ 0.28} $\uparrow$ } \\ 
&All - \task{semtr}&{\color{olive}68.17\tiny{ $\pm$ 0.15} $\uparrow$ }&{\color{olive}68.16\tiny{ $\pm$ 0.47} $\uparrow$ }&67.96\tiny{ $\pm$ 0.73} \\ 
&All - \task{com}&{\color{olive}68.67\tiny{ $\pm$ 0.37} $\uparrow$ }&67.62\tiny{ $\pm$ 0.6}&{\color{olive}67.94\tiny{ $\pm$ 0.22} $\uparrow$ } \\ 
&All - \task{frame}&{\color{olive}68.47\tiny{ $\pm$ 0.72} $\uparrow$ }&67.69\tiny{ $\pm$ 0.95}&{\color{olive}68.13\tiny{ $\pm$ 0.39} $\uparrow$ } \\ 
&All - \task{hyp}&{\color{olive}68.46\tiny{ $\pm$ 0.37} $\uparrow$ }&{\color{olive}68.32\tiny{ $\pm$ 0.18} $\uparrow$ }&{\color{olive}68.17\tiny{ $\pm$ 0.36} $\uparrow$ } \\  \hline 
\multicolumn{2}{c|}{All}&{\color{olive}68.1\tiny{ $\pm$ 0.54} $\uparrow$ }&{\color{olive}67.98\tiny{ $\pm$ 0.29} $\uparrow$ }&{\color{olive}68.02\tiny{ $\pm$ 0.21} $\uparrow$ } \\  \hline 
\multicolumn{2}{c|}{ Oracle } & 68.53\tiny{ $\pm$ 0.09}&68.22\tiny{ $\pm$ 0.61}&69.04\tiny{ $\pm$ 0.44} \\ \hline
\end{tabular}
\caption{\small F1 score tested on the task \task{supsense} in different training scenarios}\label{tMultiTasksupsensestreusle}}}
\end{table*}

\begin{table*}[t]
\parbox{.49\linewidth}{
\centering
\scriptsize{
\begin{tabular}{c|c|c|c|c}
\multicolumn{2}{c}{Trained with} & \multicolumn{3}{|c}{Tested on \task{com}} \\ \cline{3-5}
 \multicolumn{2}{c|}{} & \multidec & \tedec & \teenc \\ \hline
\multicolumn{2}{c}{ \task{com} only } & \multicolumn{3}{|c}{72.71\tiny{ $\pm$ 0.75}}\\ \hline
\parbox[t]{1mm}{\multirow{11}{*}{\rotatebox[origin = c]{90}{Pairwise}}}&\task{+upos}&72.46\tiny{ $\pm$ 0.34}&72.86\tiny{ $\pm$ 0.12}&72.09\tiny{ $\pm$ 0.36} \\ 
&\task{+xpos}&72.83\tiny{ $\pm$ 0.16}&72.87\tiny{ $\pm$ 0.56}&72.41\tiny{ $\pm$ 0.51} \\ 
&\task{+chunk}&72.44\tiny{ $\pm$ 0.11}&73.3\tiny{ $\pm$ 0.15}&72.88\tiny{ $\pm$ 0.26} \\ 
&\task{+ner}&70.93\tiny{ $\pm$ 0.73}&{\color{red}71.08\tiny{ $\pm$ 0.31} $\downarrow$ }&{\color{red}70.78\tiny{ $\pm$ 0.27} $\downarrow$ } \\ 
&\task{+mwe}&71.31\tiny{ $\pm$ 0.31}&{\color{red}70.93\tiny{ $\pm$ 0.43} $\downarrow$ }&71.36\tiny{ $\pm$ 0.42} \\ 
&\task{+sem}&72.72\tiny{ $\pm$ 0.22}&73.14\tiny{ $\pm$ 0.08}&72.25\tiny{ $\pm$ 0.07} \\ 
&\task{+semtr}&71.96\tiny{ $\pm$ 0.16}&71.74\tiny{ $\pm$ 0.46}&72.15\tiny{ $\pm$ 0.5} \\ 
&\task{+supsense}&72.24\tiny{ $\pm$ 0.27}&{\color{red}69.13\tiny{ $\pm$ 0.19} $\downarrow$ }&72.12\tiny{ $\pm$ 0.66} \\ 
&\task{+frame}&72.47\tiny{ $\pm$ 0.08}&72.89\tiny{ $\pm$ 0.22}&72.1\tiny{ $\pm$ 0.93} \\ 
&\task{+hyp}&71.82\tiny{ $\pm$ 0.97}&70.47\tiny{ $\pm$ 0.81}&72.79\tiny{ $\pm$ 0.97} \\  \cline{2-5} 
& Average&72.12&71.84&72.09 \\ \hline 
\parbox[t]{1mm}{\multirow{10}{*}{\rotatebox[origin = c]{90}{All-but-one}}}&All - \task{upos}&{\color{olive}74.42\tiny{ $\pm$ 0.24} $\uparrow$ }&{\color{olive}74.69\tiny{ $\pm$ 0.26} $\uparrow$ }&74.07\tiny{ $\pm$ 0.19} \\ 
&All - \task{xpos}&{\color{olive}74.36\tiny{ $\pm$ 0.14} $\uparrow$ }&74.26\tiny{ $\pm$ 0.64}&73.94\tiny{ $\pm$ 0.3} \\ 
&All - \task{chunk}&{\color{olive}74.2\tiny{ $\pm$ 0.13} $\uparrow$ }&{\color{olive}74.47\tiny{ $\pm$ 0.26} $\uparrow$ }&73.67\tiny{ $\pm$ 0.23} \\ 
&All - \task{ner}&{\color{olive}74.08\tiny{ $\pm$ 0.07} $\uparrow$ }&{\color{olive}74.49\tiny{ $\pm$ 0.38} $\uparrow$ }&74.16\tiny{ $\pm$ 0.48} \\ 
&All - \task{mwe}&{\color{olive}74.7\tiny{ $\pm$ 0.14} $\uparrow$ }&{\color{olive}74.49\tiny{ $\pm$ 0.13} $\uparrow$ }&{\color{olive}74.28\tiny{ $\pm$ 0.16} $\uparrow$ } \\ 
&All - \task{sem}&{\color{olive}74.31\tiny{ $\pm$ 0.1} $\uparrow$ }&74.34\tiny{ $\pm$ 0.42}&74.2\tiny{ $\pm$ 0.28} \\ 
&All - \task{semtr}&{\color{olive}74.2\tiny{ $\pm$ 0.24} $\uparrow$ }&74.36\tiny{ $\pm$ 0.36}&73.81\tiny{ $\pm$ 0.16} \\ 
&All - \task{supsense}&74.24\tiny{ $\pm$ 0.44}&{\color{olive}74.69\tiny{ $\pm$ 0.52} $\uparrow$ }&{\color{olive}74.3\tiny{ $\pm$ 0.13} $\uparrow$ } \\ 
&All - \task{frame}&{\color{olive}75.03\tiny{ $\pm$ 0.24} $\uparrow$ }&{\color{olive}74.49\tiny{ $\pm$ 0.2} $\uparrow$ }&{\color{olive}74.3\tiny{ $\pm$ 0.19} $\uparrow$ } \\ 
&All - \task{hyp}&{\color{olive}74.62\tiny{ $\pm$ 0.14} $\uparrow$ }&{\color{olive}74.4\tiny{ $\pm$ 0.06} $\uparrow$ }&73.78\tiny{ $\pm$ 0.05} \\  \hline 
\multicolumn{2}{c|}{All}&74.54\tiny{ $\pm$ 0.53}&{\color{olive}74.61\tiny{ $\pm$ 0.24} $\uparrow$ }&{\color{olive}74.61\tiny{ $\pm$ 0.32} $\uparrow$ } \\  \hline 
\multicolumn{2}{c|}{ Oracle } & {\color{blue}72.71 $\pm$ 0.75}&{\color{blue}72.71 $\pm$ 0.75}&{\color{blue}72.71 $\pm$ 0.75} \\ \hline
\end{tabular}
\caption{\small F1 score tested on the task \task{com} in different training scenarios}\label{tMultiTaskcombroadcast1}}}
\parbox{.49\linewidth}{
\centering
\scriptsize{
\begin{tabular}{c|c|c|c|c}
\multicolumn{2}{c}{Trained with} & \multicolumn{3}{|c}{Tested on \task{frame}} \\ \cline{3-5}
 \multicolumn{2}{c|}{} & \multidec & \tedec & \teenc \\ \hline
\multicolumn{2}{c}{ \task{frame} only } & \multicolumn{3}{|c}{62.04\tiny{ $\pm$ 0.74}}\\ \hline
\parbox[t]{1mm}{\multirow{11}{*}{\rotatebox[origin = c]{90}{Pairwise}}}&\task{+upos}&62.14\tiny{ $\pm$ 0.35}&61.54\tiny{ $\pm$ 0.53}&62.27\tiny{ $\pm$ 0.33} \\ 
&\task{+xpos}&60.77\tiny{ $\pm$ 0.39}&61.44\tiny{ $\pm$ 0.06}&61.62\tiny{ $\pm$ 1.01} \\ 
&\task{+chunk}&62.67\tiny{ $\pm$ 0.47}&61.39\tiny{ $\pm$ 0.78}&62.98\tiny{ $\pm$ 0.5} \\ 
&\task{+ner}&62.39\tiny{ $\pm$ 0.37}&{\color{red}59.25\tiny{ $\pm$ 0.52} $\downarrow$ }&63.02\tiny{ $\pm$ 0.39} \\ 
&\task{+mwe}&61.75\tiny{ $\pm$ 0.21}&56.77\tiny{ $\pm$ 2.79}&60.61\tiny{ $\pm$ 0.91} \\ 
&\task{+sem}&61.74\tiny{ $\pm$ 0.27}&{\color{red}60.09\tiny{ $\pm$ 0.48} $\downarrow$ }&62.17\tiny{ $\pm$ 0.36} \\ 
&\task{+semtr}&62.03\tiny{ $\pm$ 0.41}&59.77\tiny{ $\pm$ 0.81}&62.79\tiny{ $\pm$ 0.19} \\ 
&\task{+supsense}&61.94\tiny{ $\pm$ 0.43}&{\color{red}55.68\tiny{ $\pm$ 0.61} $\downarrow$ }&61.96\tiny{ $\pm$ 0.18} \\ 
&\task{+com}&{\color{red}56.52\tiny{ $\pm$ 0.27} $\downarrow$ }&{\color{red}55.25\tiny{ $\pm$ 2.29} $\downarrow$ }&57.65\tiny{ $\pm$ 2.42} \\ 
&\task{+hyp}&61.02\tiny{ $\pm$ 0.62}&{\color{red}55.35\tiny{ $\pm$ 0.5} $\downarrow$ }&61.14\tiny{ $\pm$ 1.77} \\  \cline{2-5} 
& Average&61.3&58.65&61.62 \\ \hline 
\parbox[t]{1mm}{\multirow{10}{*}{\rotatebox[origin = c]{90}{All-but-one}}}&All - \task{upos}&{\color{red}58.47\tiny{ $\pm$ 1.0} $\downarrow$ }&{\color{red}58.32\tiny{ $\pm$ 0.35} $\downarrow$ }&{\color{red}60.51\tiny{ $\pm$ 0.1} $\downarrow$ } \\ 
&All - \task{xpos}&{\color{red}60.16\tiny{ $\pm$ 0.42} $\downarrow$ }&{\color{red}58.31\tiny{ $\pm$ 0.8} $\downarrow$ }&60.13\tiny{ $\pm$ 1.38} \\ 
&All - \task{chunk}&60.01\tiny{ $\pm$ 0.65}&{\color{red}58.73\tiny{ $\pm$ 0.68} $\downarrow$ }&61.73\tiny{ $\pm$ 0.48} \\ 
&All - \task{ner}&{\color{red}59.17\tiny{ $\pm$ 0.27} $\downarrow$ }&{\color{red}58.19\tiny{ $\pm$ 0.89} $\downarrow$ }&{\color{red}59.96\tiny{ $\pm$ 0.52} $\downarrow$ } \\ 
&All - \task{mwe}&{\color{red}59.23\tiny{ $\pm$ 0.33} $\downarrow$ }&{\color{red}57.6\tiny{ $\pm$ 0.82} $\downarrow$ }&61.51\tiny{ $\pm$ 0.43} \\ 
&All - \task{sem}&{\color{red}58.73\tiny{ $\pm$ 0.67} $\downarrow$ }&{\color{red}59.08\tiny{ $\pm$ 0.84} $\downarrow$ }&61.76\tiny{ $\pm$ 0.52} \\ 
&All - \task{semtr}&{\color{red}59.49\tiny{ $\pm$ 0.79} $\downarrow$ }&{\color{red}58.85\tiny{ $\pm$ 0.51} $\downarrow$ }&61.31\tiny{ $\pm$ 1.16} \\ 
&All - \task{supsense}&{\color{red}59.23\tiny{ $\pm$ 0.64} $\downarrow$ }&{\color{red}58.28\tiny{ $\pm$ 0.19} $\downarrow$ }&59.98\tiny{ $\pm$ 1.23} \\ 
&All - \task{com}&62.37\tiny{ $\pm$ 0.37}&60.72\tiny{ $\pm$ 0.73}&63.55\tiny{ $\pm$ 0.31} \\ 
&All - \task{hyp}&{\color{red}59.69\tiny{ $\pm$ 0.41} $\downarrow$ }&{\color{red}58.55\tiny{ $\pm$ 0.29} $\downarrow$ }&61.91\tiny{ $\pm$ 0.59} \\  \hline 
\multicolumn{2}{c|}{All}&59.71\tiny{ $\pm$ 0.85}&{\color{red}58.14\tiny{ $\pm$ 0.23} $\downarrow$ }&61.83\tiny{ $\pm$ 0.98} \\  \hline 
\multicolumn{2}{c|}{ Oracle } & {\color{blue}62.04 $\pm$ 0.74}&{\color{blue}62.04 $\pm$ 0.74}&{\color{blue}62.04 $\pm$ 0.74} \\ \hline
\end{tabular}
\caption{\small F1 score tested on the task \task{frame} in different training scenarios}\label{tMultiTaskframefnt}}}
\parbox{.49\linewidth}{
\centering
\scriptsize{
\begin{tabular}{c|c|c|c|c}
\multicolumn{2}{c}{Trained with} & \multicolumn{3}{|c}{Tested on \task{hyp}} \\ \cline{3-5}
 \multicolumn{2}{c|}{} & \multidec & \tedec & \teenc \\ \hline
\multicolumn{2}{c}{ \task{hyp} only } & \multicolumn{3}{|c}{46.73\tiny{ $\pm$ 0.55}}\\ \hline
\parbox[t]{1mm}{\multirow{11}{*}{\rotatebox[origin = c]{90}{Pairwise}}}&\task{+upos}&48.02\tiny{ $\pm$ 0.31}&{\color{olive}49.36\tiny{ $\pm$ 0.36} $\uparrow$ }&48.27\tiny{ $\pm$ 0.68} \\ 
&\task{+xpos}&{\color{olive}48.81\tiny{ $\pm$ 0.36} $\uparrow$ }&{\color{olive}49.23\tiny{ $\pm$ 0.55} $\uparrow$ }&{\color{olive}48.06\tiny{ $\pm$ 0.02} $\uparrow$ } \\ 
&\task{+chunk}&{\color{olive}47.85\tiny{ $\pm$ 0.2} $\uparrow$ }&{\color{olive}48.43\tiny{ $\pm$ 0.3} $\uparrow$ }&47.13\tiny{ $\pm$ 0.35} \\ 
&\task{+ner}&47.9\tiny{ $\pm$ 0.67}&48.24\tiny{ $\pm$ 0.65}&48.64\tiny{ $\pm$ 1.17} \\ 
&\task{+mwe}&47.32\tiny{ $\pm$ 0.29}&45.83\tiny{ $\pm$ 0.46}&46.71\tiny{ $\pm$ 0.64} \\ 
&\task{+sem}&{\color{olive}48.15\tiny{ $\pm$ 0.21} $\uparrow$ }&47.95\tiny{ $\pm$ 0.75}&47.12\tiny{ $\pm$ 0.43} \\ 
&\task{+semtr}&47.74\tiny{ $\pm$ 0.57}&46.96\tiny{ $\pm$ 0.85}&46.1\tiny{ $\pm$ 0.11} \\ 
&\task{+supsense}&{\color{olive}49.23\tiny{ $\pm$ 0.13} $\uparrow$ }&47.29\tiny{ $\pm$ 0.41}&47.24\tiny{ $\pm$ 0.43} \\ 
&\task{+com}&47.41\tiny{ $\pm$ 1.18}&45.24\tiny{ $\pm$ 0.46}&47.81\tiny{ $\pm$ 0.8} \\ 
&\task{+frame}&47.5\tiny{ $\pm$ 0.46}&46.0\tiny{ $\pm$ 0.53}&46.66\tiny{ $\pm$ 0.54} \\  \cline{2-5} 
& Average&47.99&47.45&47.37 \\ \hline 
\parbox[t]{1mm}{\multirow{10}{*}{\rotatebox[origin = c]{90}{All-but-one}}}&All - \task{upos}&{\color{olive}51.13\tiny{ $\pm$ 0.94} $\uparrow$ }&{\color{olive}50.83\tiny{ $\pm$ 0.65} $\uparrow$ }&{\color{olive}50.23\tiny{ $\pm$ 0.73} $\uparrow$ } \\ 
&All - \task{xpos}&{\color{olive}51.65\tiny{ $\pm$ 0.63} $\uparrow$ }&{\color{olive}50.6\tiny{ $\pm$ 0.44} $\uparrow$ }&{\color{olive}50.39\tiny{ $\pm$ 1.17} $\uparrow$ } \\ 
&All - \task{chunk}&{\color{olive}50.27\tiny{ $\pm$ 0.76} $\uparrow$ }&{\color{olive}51.1\tiny{ $\pm$ 0.28} $\uparrow$ }&{\color{olive}50.18\tiny{ $\pm$ 0.81} $\uparrow$ } \\ 
&All - \task{ner}&{\color{olive}50.86\tiny{ $\pm$ 0.87} $\uparrow$ }&{\color{olive}50.44\tiny{ $\pm$ 0.39} $\uparrow$ }&{\color{olive}49.95\tiny{ $\pm$ 0.38} $\uparrow$ } \\ 
&All - \task{mwe}&{\color{olive}50.83\tiny{ $\pm$ 0.61} $\uparrow$ }&{\color{olive}50.5\tiny{ $\pm$ 0.9} $\uparrow$ }&{\color{olive}49.81\tiny{ $\pm$ 0.44} $\uparrow$ } \\ 
&All - \task{sem}&{\color{olive}50.93\tiny{ $\pm$ 0.27} $\uparrow$ }&{\color{olive}50.48\tiny{ $\pm$ 0.53} $\uparrow$ }&{\color{olive}50.15\tiny{ $\pm$ 0.11} $\uparrow$ } \\ 
&All - \task{semtr}&{\color{olive}51.27\tiny{ $\pm$ 0.5} $\uparrow$ }&{\color{olive}51.5\tiny{ $\pm$ 0.46} $\uparrow$ }&{\color{olive}51.72\tiny{ $\pm$ 0.15} $\uparrow$ } \\ 
&All - \task{supsense}&{\color{olive}50.86\tiny{ $\pm$ 1.85} $\uparrow$ }&{\color{olive}51.96\tiny{ $\pm$ 0.29} $\uparrow$ }&{\color{olive}50.01\tiny{ $\pm$ 1.13} $\uparrow$ } \\ 
&All - \task{com}&{\color{olive}50.28\tiny{ $\pm$ 1.02} $\uparrow$ }&{\color{olive}50.0\tiny{ $\pm$ 0.11} $\uparrow$ }&{\color{olive}48.77\tiny{ $\pm$ 0.54} $\uparrow$ } \\ 
&All - \task{frame}&{\color{olive}50.89\tiny{ $\pm$ 0.64} $\uparrow$ }&{\color{olive}51.23\tiny{ $\pm$ 1.01} $\uparrow$ }&{\color{olive}50.35\tiny{ $\pm$ 0.68} $\uparrow$ } \\  \hline 
\multicolumn{2}{c|}{All}&{\color{olive}51.41\tiny{ $\pm$ 0.25} $\uparrow$ }&{\color{olive}51.31\tiny{ $\pm$ 0.55} $\uparrow$ }&{\color{olive}49.5\tiny{ $\pm$ 0.05} $\uparrow$ } \\  \hline 
\multicolumn{2}{c|}{ Oracle } & 50.0\tiny{ $\pm$ 0.42}&50.15\tiny{ $\pm$ 0.25}&48.06\tiny{ $\pm$ 0.02} \\ \hline
\end{tabular}
\caption{\small F1 score tested on the task \task{hyp} in different training scenarios}\label{tMultiTaskhyphyp}}}
\end{table*}

\end{landscape}

\end{document}